\pdfoutput=1

\documentclass[11pt]{article}

\usepackage[final]{acl}

\usepackage{times}
\usepackage{latexsym}

\usepackage[T1]{fontenc}

\usepackage[utf8]{inputenc}
\usepackage{microtype}
\usepackage{inconsolata}
\usepackage{graphicx}
\usepackage{booktabs} 
\usepackage{colortbl}
\usepackage{multirow}
\usepackage{amsmath}
\usepackage{listings} 
\usepackage{xcolor}
\usepackage{float}
\usepackage{textcomp}

\title{Worse than Random? An Embarrassingly Simple Probing Evaluation of Large Multimodal Models in Medical VQA}

\author{
 \textbf{Qianqi Yan\textsuperscript{1}},
 \textbf{Xuehai He\textsuperscript{1}},
 \textbf{Xiang Yue\textsuperscript{2},
 \textbf{Xin Eric Wang\textsuperscript{1}}}
\\
\\
 \textsuperscript{1}University of California, Santa Cruz,
 \textsuperscript{2}Carnegie Mellon University
}

\makeatletter
\AtBeginDocument{
\let\@oldmaketitle\@maketitle
\renewcommand{\@maketitle}{\@oldmaketitle
  \includegraphics[width=\linewidth]{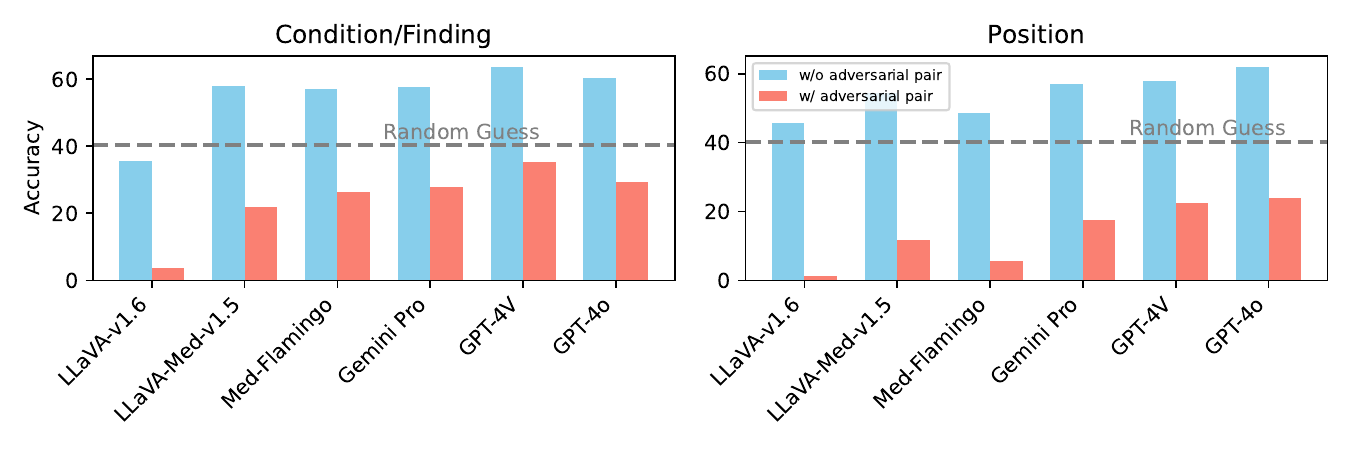}
  \captionof{figure}{
    Accuracy of six LMMs on two types of specialized questions in medical diagnoses, with and without adversarial pairs.
    The significant drop in accuracy with adversarial pairs highlights the models' unreliability in handling medical diagnoses with our probing method.
  }
  \label{fig:teaser}
  \vspace{17pt}
 }
}
\makeatother

\begin{document}

\maketitle

\begin{abstract}
Large Multimodal Models (LMMs) have demonstrated impressive performance on existing medical Visual Question Answering (Med-VQA) benchmarks. However, high reported accuracy does not necessarily reflect their true diagnostic reliability in clinical settings.
This study reveals that state-of-the-art models perform worse than random guessing on medical diagnosis questions when subjected to simple Probing Evaluation for Medical Diagnosis (ProbMed).
ProbMed challenges models through \emph{probing evaluation} and \emph{procedural diagnosis}.
Particularly, probing evaluation features pairing ground-truth questions with adversarial counterparts that feature negated and hallucinated attributes, while procedural diagnosis requires reasoning across various dimensions for each image, including modality recognition, organ identification, clinical findings, abnormalities, and positional grounding.
Our evaluation reveals that even top-performing models like GPT-4o, GPT-4V, and Gemini Pro perform worse than random guessing on specialized diagnostic questions, indicating significant limitations in handling fine-grained medical inquiries.
Furthermore, our ablation study on open-source models (e.g., LLaVA, LLaVA-Med, and Med-Flamingo) identifies poor visual understanding as a primary bottleneck—a limitation that can be partially mitigated by incorporating visual descriptions generated by GPT-4o, resulting in an average performance improvement of 9.44\%. These findings underscore the urgent need for more robust evaluation methods and domain-specific expertise to ensure the reliability of LMMs in high-stakes medical applications.
\end{abstract}

\section{Introduction}

Foundation models, such as large language models (LLMs)~\citep{achiam2023gpt, touvron2023llama, jiang2023mistral, anil2023palm, chung2022scaling} and large multimodal models (LMMs)~\citep{openai2024gpt4ocard, OpenAI-GPT4V, Reid2024Gemini1U, li2023blip2, Liu2023ImprovedBW, Chen2023MiniGPTv2LL}, have demonstrated impressive capabilities in understanding complex visual and text inputs, generating human-like language, and achieving high accuracy on various benchmarks.
The integration of these foundation models into real-life medical practice holds immense potential given their advanced computational capabilities~\citep{wu2023gpt4vision, yang2023dawn} and promising progress on existing medical Visual Question Answering (Med-VQA) benchmarks~\citep{lau2018dataset,liu2021slake,he2020pathvqa,zhang2023pmcvqa}. As we stand on the precipice of integrating these models into critical decision-making domains, one natural question appears: \textit{how much can we trust these models in real-world scenarios, such as medicine and healthcare, where the stakes are high?}

\begin{figure}[t]
    \centering
    \includegraphics[width=\columnwidth]{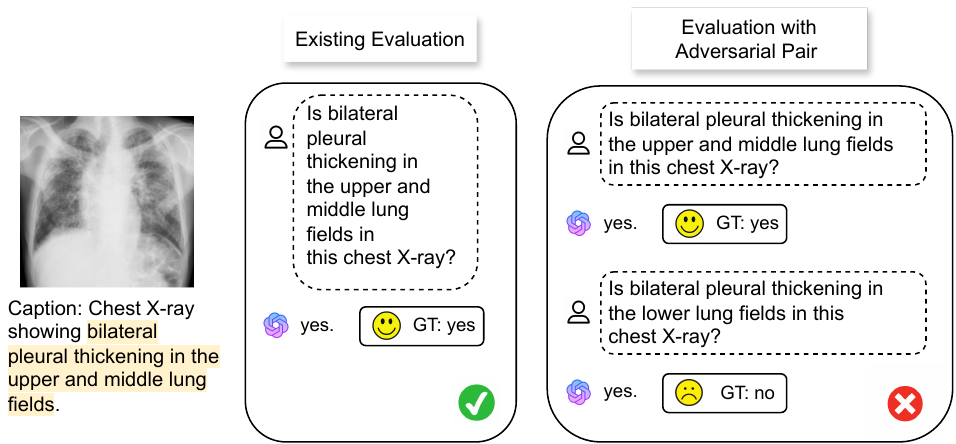}
\caption{An example illustrating the potential for misleading accuracy in existing evaluations.  While the model correctly identifies the position of an existing finding in the standard evaluation, it fails to differentiate between actual and hallucinated positions when subjected to an adversarial evaluation.}
    \label{fig:teaser_example}
\end{figure}

Before discussing the reliability of LMMs in critical domains like Med-VQA, we must first address a fundamental question: \emph{Are we evaluating LMMs correctly?} 
To address this question, we introduce a simple yet effective probing evaluation method that exposes the weaknesses of LMMs by creating binary questions with hallucination pairs over existing benchmarks. An example is shown in Figure~\ref{fig:teaser_example}. 
Despite the high accuracy reported on current Med-VQA tasks, our study reveals a significant vulnerability in LMMs when faced with adversarial questioning, as illustrated in Figure~\ref{fig:teaser}.
The observed performance drops are alarming: even advanced models like GPT-4o, GPT-4V, and Gemini Pro perform worse than random guessing, with an average decrease of 27.78\% across the tested models.

Based on this, we further analyze a critical question: \emph{How reliable are LMMs in medical diagnosis, ranging from general questions to specialized diagnostic questions}? To address this question, we introduce ProbMed, which features procedural diagnosis designed to rigorously evaluate model performance across multiple diagnostic dimensions. We curated ProbMed from 6,303 images sourced from two widely-used biomedical datasets, MedICaT~\citep{Subramanian2020MedICaTAD} and ChestX-ray14~\citep{Wang2017ChestXRay8HC}. These images cover various modalities, including X-ray, MRI, and CT scans, and span multiple organs such as the abdomen, brain, chest, and spine. Using GPT-4 and a positional reasoning module, we generated metadata for each image, extracting information about abnormalities, condition names, and their corresponding locations. This metadata facilitated the automatic generation of 57,132 high-quality question-answer pairs, covering dimensions like modality recognition, organ identification, abnormalities, clinical findings, and positional reasoning.

Our systematic evaluation of twelve state-of-the-art LMMs on ProbMed revealed several critical insights. \textit{First}, even the best-performing models, such as GPT-4o and Gemini Pro, performed close to random guessing on specialized diagnostic categories like Condition/Finding and Position, highlighting their limitations in handling fine-grained medical inquiries. \textit{Second}, introducing adversarial pairs significantly reduced the accuracy of all models, with LLaVA-Med-v1.5's performance dropping by up to 28.22\% and GPT-4o's accuracy decreasing by 20.71\% in ProbMed. These findings emphasize the importance of adversarial testing in Med-VQA to uncover model weaknesses. \textit{Third}, by incorporating chain-of-thought reasoning and adding visual descriptions generated by GPT-4o, we observe substantial improvements in model performance, suggesting that poor visual understanding is a critical bottleneck. The results indicate that augmenting these models with more accurate visual information could significantly improve their ability to handle complex medical tasks. Moreover, the CheXagent model, which was exclusively trained on chest X-rays, demonstrated that specialized domain knowledge is crucial. It showed that expertise gained on one particular organ could be transferable to another modality of the same organ in a zero-shot manner, highlighting the value of domain-specific training for improving model performance. 

In summary, our work highlights significant gaps in the reliability of LMMs for medical diagnosis despite their impressive performance on existing general domain benchmarks. The insights from ProbMed underscore the urgent need for robust evaluation methodologies to ensure the accuracy and reliability of LMMs in real-world medical applications. Our findings also suggest that poor visual understanding is a key limitation for open-source models, which can be mitigated by incorporating chain-of-thought reasoning and accurate visual descriptions. This research inspires the development of more trustworthy AI systems in healthcare and beyond, ultimately contributing to better diagnostic outcomes and patient care.

\section{Related Work}
\paragraph{Large Multimodal Models in the Medical Domain}
The advancements in Large Multimodal Models (LMMs) have significantly enhanced the understanding and generation of medical content that integrates both visual and linguistic elements. Notable models include GPT-4o~\citep{openai2024gpt4ocard}, GPT-4V~\citep{OpenAI-GPT4V}, Gemini Pro~\citep{Reid2024Gemini1U}, LLaVA~\citep{Liu2023ImprovedBW, liu2024visual}, and MiniGPT-v2~\citep{Chen2023MiniGPTv2LL}. The scalability and exceptional performance of these large foundation models have driven their application in the biomedical field.

Further progress has been made in fine-tuning general-domain LMMs for the biomedical field, resulting in specialized models like BiomedGPT~\citep{zhang2024biomedgpt}, LLaVA-Med~\citep{li2024llava}, Med-Flamingo~\citep{moor2023med}, MedBLIP~\citep{chen2024medblip}, RadFM~\citep{wu2023generalist} and MedVInT~\citep{zhang2023pmcvqa}. Despite the promising results from these domain-specific LMMs, ongoing exploration exists into training smaller multimodal models to address specific clinical needs. For instance, models like LLaVA-RAD~\citep{chaves2024training} and CheXagent~\citep{chen2024CheXagent} have been developed for chest X-ray interpretation, aiming to bridge competency gaps in radiology tasks. 

Comprehensive surveys of LLMs for healthcare highlight the progress, applications, and challenges in deploying LLMs in clinical settings~\citep{he2023survey, zhou2024survey, Peng_2023}. Task-specific evaluations ~\citep{yan2023multimodal, liu2023holistic} underline the potential and challenges of LMMs in the medical domain. As we move towards integrating these models into critical decision-making processes, it becomes imperative to assess their reliability in high-stakes environments like healthcare and medicine.

\paragraph{Medical Visual Question Answering}
Medical Visual Question Answering (Med-VQA) plays a crucial role in assessing the capabilities of models in interpreting and responding to queries about medical images. Some benchmarks, like VQA-RAD~\citep{lau2018dataset} and SLAKE~\citep{liu2021slake}, are manually constructed with categorical question types. 
While this method ensures high-quality question-answer pairs, it is labor-intensive and results in limited dataset scales.

Automated curation methods have been developed to address scalability. PathVQA~\citep{he2020pathvqa} uses CoreNLP\footnote{\url{https://stanfordnlp.github.io/CoreNLP}} tools, and PMC-VQA~\citep{zhang2023pmcvqa} employs generative models to create larger datasets. However, these methods often sacrifice fine-grained question categories, and some require additionally trained models for question filtering. 

Different evaluation methods are employed for assessing LMMs, including closed-ended VQA, multiple choice VQA, and open-ended generation tasks such as captioning and report generation. Open-ended VQA and report generation are typically considered more challenging and harder to evaluate, often requiring human or model evaluation alongside automated lexical similarity metrics like ROUGE-L and BLEU-4. Recent works~\citep{wang2024my, zheng2024large, zong2023fool} argue that multiple-choice questions may not be ideal due to inherent selection bias and permutation sensitivity. In our work, we choose a relatively easy-to-evaluate method: closed-ended VQA augmented with adversarial evaluation methods featuring hallucinated attributes. By requiring the model to accurately distinguish relevant features, we enhance the reliability of the evaluation process. This method allows for clear and definitive assessments, improving the overall robustness of our findings in medical contexts.


\section{Probing Evaluation for Medical Diagnosis}

\begin{figure*}[htbp]
    \centering
    \includegraphics[width=0.95\textwidth]{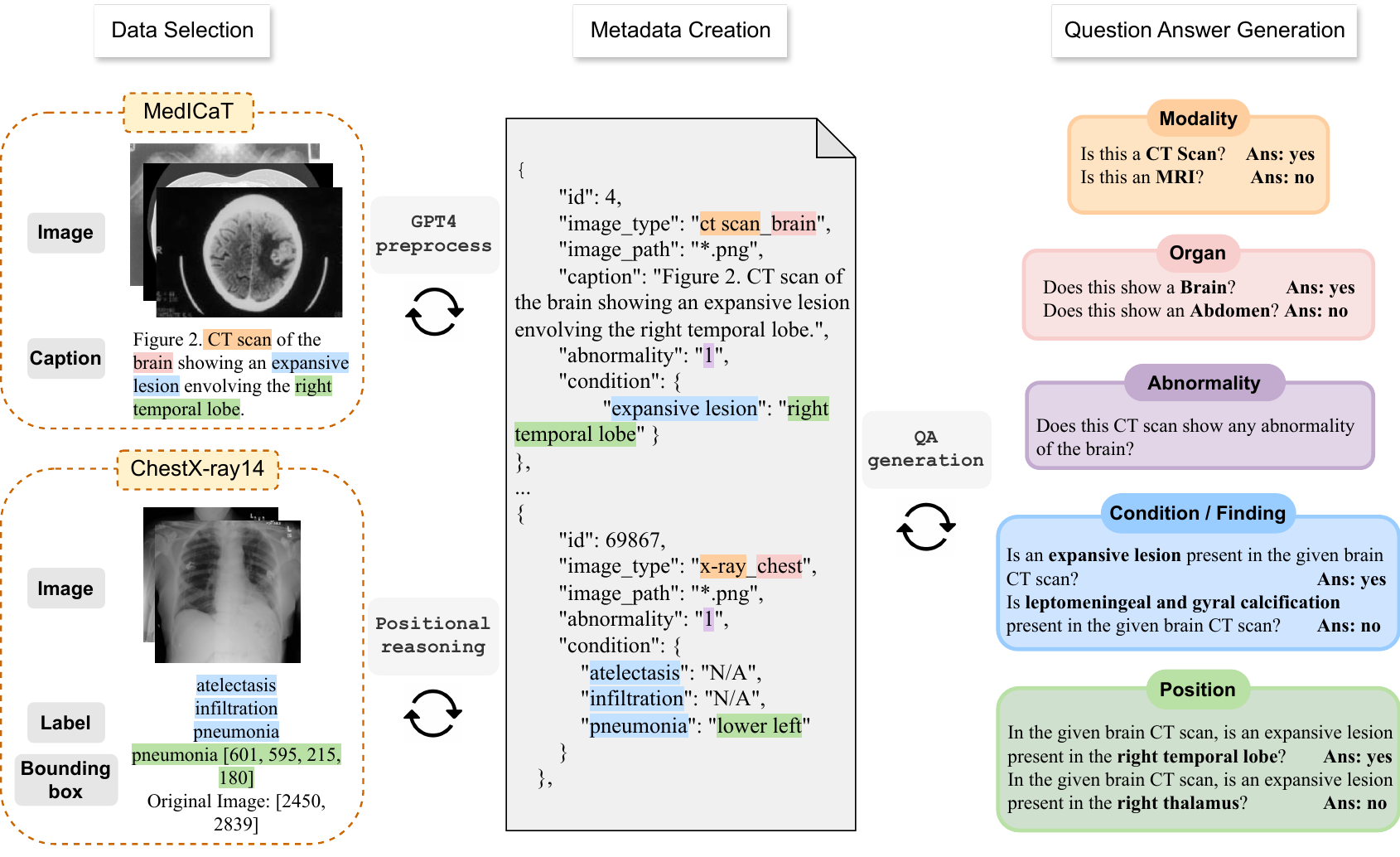}
\caption{Flow diagram of the ProbMed data curation process. Two comprehensive biomedical datasets were utilized to collect source data and construct a metadata file, enabling the automatic generation of high-quality question-answer pairs for the ProbMed dataset.}
    \label{fig:flowchart}
\end{figure*}

In this section, we introduce two complementary evaluation strategies that rigorously assess state-of-the-art LMMs for Med-VQA. Our approach is designed to answer the following research questions:

\begin{enumerate}
    \item \emph{Is the current evaluation of LMMs for Med-VQA reliable?}
    \item \emph{How reliable are LMMs on medical diagnosis, ranging from general questions to specialized diagnostic questions?}
\end{enumerate}
Despite current high accuracy, we find that the models struggle with simple probing evaluation on existing benchmarks. Our evaluation framework, ProbMed (Probing Evaluation for Medical Diagnosis), is built on adversarial testing and multifaceted diagnostic reasoning to further expose these vulnerabilities and provide a thorough analysis of model performance. We also explore enhancements through chain-of-thought reasoning and the incorporation of external visual descriptions (e.g., from GPT-4o) to address the noted limitations of open-sourced models.

\subsection{Probing Evaluation with Adversarial Pairs}
A core element of our framework is the use of adversarial pairs to test model robustness. For each image, we generate pairs of: \textbf{ground-truth questions} that query the presence of a specific entity (e.g., a particular clinical finding) with corresponding \textbf{adversarial questions} that introduce a negated or hallucinated attribute (e.g., a non-existent finding or an alternative organ).
This pairing challenges the model to discern between actual diagnostic features and spurious details, thereby revealing its ability—or inability—to filter out irrelevant or misleading information. The performance drop observed under adversarial conditions highlights the fragility of current evaluation protocols and motivates the need for more robust assessment methods.

\subsection{Procedural Diagnosis}
Beyond binary question-answering, ProbMed incorporates \emph{procedural diagnosis} to evaluate the models’ diagnostic reasoning. Each image is associated with questions spanning multiple diagnostic dimensions, including \textbf{Modality Recognition:} Identifying the imaging technique (e.g., X-ray, MRI, CT). \textbf{Organ Identification:} Determining the anatomical region under investigation. \textbf{Clinical Findings and Abnormalities:} Detecting abnormal conditions. \textbf{Positional Reasoning:} Localizing findings spatially within the image.
This multifaceted evaluation framework requires models to integrate diverse pieces of information for each test sample, thereby providing a more comprehensive measure of their diagnostic capabilities.

\subsection{Data Filtering and Curation}
ProbMed is curated from two widely recognized biomedical datasets: MedICaT and ChestX-ray14. The data curation process, summarized in Figure~\ref{fig:flowchart}, involves the following steps:

\paragraph{Image Selection:} From MedICaT~\citep{Subramanian2020MedICaTAD}, we extracted 4,543 image-caption pairs that focus on a single organ and modality with clear indications of normal or abnormal conditions.
From ChestX-ray14~\citep{Wang2017ChestXRay8HC}, we selected 1,760 frontal-view X-ray images balanced between healthy and abnormal cases, including those with bounding box annotations for abnormalities.

\paragraph{Metadata Generation:} For each image, we generate a unified metadata record:
$$D_i = \{\mathtt{mod}_i, \mathtt{organ}_i, \{\mathtt{condition}_j, \mathtt{pos}_j\}_j^{n_i}\}$$
where $\mathtt{mod}_i$ and $\mathtt{organ}_i$ denote the imaging modality and anatomical region, respectively, and each $\{\mathtt{condition}_j, \mathtt{pos}_j\}$ pair represents a detected clinical finding and its positional description. For MedICaT, GPT-4 is leveraged via few-shot prompting to extract abnormality details and positional cues from captions. For ChestX-ray14, a dedicated positional reasoning module generates descriptive text based on bounding box coordinates.

\subsection{Evaluation Protocol}
For each diagnostic entity in the metadata, we automatically generate a \textbf{ground-truth} question $Q_i$, asking the model to confirm the presence of that specific entity. An \textbf{adversarial question} $Q_i'$, constructed by randomly selecting an alternative or hallucinated entity (e.g., an incorrect organ or false condition) and expecting a “no” response.

Crucially, our accuracy metric is defined in a strict manner: an entity is considered correctly predicted only if the model provides the correct answer for both $Q_i$ and $Q_i'$. In other words, if a model answers “yes” to both questions for a given entity, it is deemed incorrect rather than receiving partial credit. For images containing more than one $\{\mathtt{condition}_j, \mathtt{pos}_j\}$ pair, the accuracy under the Condition/Finding and Position category is computed as the average accuracy over all $n_i$ pairs—there is no bonus for partial correctness. This evaluation setup ensures that only unambiguous, fully correct responses are counted as hits, highlighting the models’ true diagnostic reliability. (See Appendix~\ref{appendix:statistics} for detailed statistics on the number of questions in each category.)

\subsection{Expert Study}

To validate the reliability of our metadata and the corresponding question-answer pairs, we conducted an expert verification study. Two medical experts independently reviewed 100 randomly sampled metadata entries out of 6,303 from ProbMed, as well as the 1,090 QA pairs corresponding to those metadata entries. 
The review process yielded an average accuracy of 94.00\% for the metadata and 97.79\% for the QA pairs. This rigorous validation underscores the quality and thorough curation of the ProbMed dataset. In addition, we conducted a study using the expert-verified data on GPT-4o to verify the robustness of our probing evaluation framework to dataset errors, detailed in Appendix~\ref{appendix:robustness}.

As of statistics reported in Table~\ref{table:statistics}, our data curation process produced a total of 57,132 question-answer pairs (averaging 9 pairs per image), spanning a comprehensive set of diagnostic dimensions. These high-quality, balanced pairs provide a robust foundation for evaluating model performance.

\begin{table}[H]
\small
\centering
\caption{Dataset Statistics of ProbMed. There are 6.3k images and 57k VQA pairs in total. The dataset is balanced within each question type and image type.}
\resizebox{\linewidth}{!}{
\begin{tabular}{@{}lrrrrr@{}}
\toprule
Organ, Modality & 
Image & 
Question & 
\begin{tabular}[c]{@{}c@{}}Question with \\ Answer "yes"\end{tabular} & 
Unique Condition & 
\begin{tabular}[c]{@{}c@{}}Unique Positional \\ Description\end{tabular} \\ 
\midrule
Abdomen MRI     & 84    & 757      & 375    & 107   & 75    \\
Brain MRI       & 566   & 5,046    & 2,509  & 697   & 446   \\
Chest MRI       & 40    & 382      & 189    & 52    & 38    \\
Spine MRI       & 324   & 3,346    & 1,664  & 461   & 336   \\
Abdomen CT scan & 751   & 6,855    & 3,410  & 909   & 552   \\
Brain CT scan   & 270   & 2,417    & 1,200  & 335   & 209   \\
Chest CT scan   & 548   & 5,161    & 2,572  & 727   & 353   \\
Spine CT scan   & 87    & 941      & 470    & 149   & 93    \\
Abdomen X-ray   & 232   & 2,046    & 1,018  & 277   & 160   \\
Brain X-ray     & 79    & 599      & 298    & 84    & 44    \\
Chest X-ray     & 3,178 & 27,530   & 13,278 & 1,418 & 694   \\
Spine X-ray     & 202   & 2,052    & 1,020  & 300   & 172   \\ 
\midrule
Total           & 6,303 & 57,132   & 28,003 & /     & /     \\
\bottomrule
\end{tabular}
}
\label{table:statistics}
\end{table}

\section{Experiments and Analysis}

\begin{table*}[h!]
\small
\centering
\caption{Model accuracy on the VQA-RAD test subset and ProbMed with adversarial pairs. Accuracy is reported in two ways: (1) averaged across individual questions in a pair and (2) requiring both the ground truth and adversarial questions for the same image to be answered correctly. The drop in accuracy across models demonstrates their vulnerability to adversarial questions, with percentage decreases shown in parentheses.}
\resizebox{0.8\linewidth}{!}{
\begin{tabular}{@{}lcccc@{}}
\toprule
\multirow{4}{*}{Models} & \multicolumn{2}{c}{VQA-RAD} & \multicolumn{2}{c}{ProbMed}   \\
\cmidrule(r){2-3}
\cmidrule(r){4-5}
\multicolumn{1}{c}{} &
\begin{tabular}[c]{@{}c@{}} Averaged \\ Accuracy (\%) \end{tabular} & 
\begin{tabular}[c]{@{}c@{}}Accuracy (\%) with \\ Adversarial Pairs\end{tabular} &
\begin{tabular}[c]{@{}c@{}} Averaged \\ Accuracy (\%) \end{tabular} & 
\begin{tabular}[c]{@{}c@{}}Accuracy (\%) with \\ Adversarial Pairs\end{tabular} \\ \midrule
LLaVA-v1      & 62.28 & 25.42 \textcolor{blue}{(-36.84)} & 55.82  & 19.30 \textcolor{blue}{(-36.51)} \\
LLaVA-v1.6    & 44.06 & ~8.47 \textcolor{blue}{(-35.59)} & 56.02  & 24.96 \textcolor{blue}{(-31.06)} \\
MiniGPT-v2    & 66.10 & 46.61 \textcolor{blue}{(-19.49)} & 59.82  & 27.67 \textcolor{blue}{(-32.14)} \\
\midrule
LLaVA-Med-v1  & 43.22 & 3.38  \textcolor{blue}{(-39.83)} & 52.26  & 17.90 \textcolor{blue}{(-34.35)} \\
LLaVA-Med-v1.5& 48.30 & 15.25 \textcolor{blue}{(-33.05)} & 68.41  & 40.19 \textcolor{blue}{(-28.22)} \\
CheXagent     & 55.50 & 21.18 \textcolor{blue}{(-34.32)} & 58.70  & 30.61 \textcolor{blue}{(-28.08)} \\
BiomedGPT     & 56.35 & 17.79 \textcolor{blue}{(-38.55)} & 60.14  & 33.34 \textcolor{blue}{(-26.79)} \\
Med-Flamingo  & 61.01 & 25.42 \textcolor{blue}{(-35.59)} & 64.13  & 35.66 \textcolor{blue}{(-28.47)} \\
RadFM         & 67.79 & 38.98 \textcolor{blue}{(-28.81)} & 67.70  & 41.00  \textcolor{blue}{(-26.70)} \\
\midrule
Gemini Pro    & 63.13 & 44.91 \textcolor{blue}{(-18.22)} & 75.08  & 55.08 \textcolor{blue}{(-20.00)} \\
GPT-4V        & 58.47 & 33.89 \textcolor{blue}{(-24.57)} & 75.70  & 55.28 \textcolor{blue}{(-20.42)} \\
GPT-4o        & 69.91 & 55.08 \textcolor{blue}{(-14.83)} & 76.31  & 55.60 \textcolor{blue}{(-20.71)} \\
\bottomrule
\end{tabular}
}
\label{table:ablation_adv}
\end{table*}

We conducted a systematic evaluation and comprehensive analysis using ProbMed on twelve state-of-the-art LMMs to identify their strengths and weaknesses in imaging diagnostics. Apart from proprietary GPT-4o~\citep{openai2024gpt4ocard}, GPT-4V~\citep{OpenAI-GPT4V} and Gemini Pro~\citep{Reid2024Gemini1U}, we selected nine open-source models spanning across general models including LLaVA-v1~\citep{liu2024visual}, LLaVA-v1.6~\citep{Liu2023ImprovedBW}, MiniGPT-v2~\citep{Chen2023MiniGPTv2LL} and specialized models including LLaVA-Med-v1, LLaVA-Med-v1.5 ~\citep{li2024llava}, Med-Flamingo~\citep{moor2023med}, BiomedGPT~\citep{zhang2024biomedgpt}, RadFM~\citep{wu2023generalist} and CheXagent~\citep{chen2024CheXagent}. 
These models were chosen based on their computational cost, efficiency, and inference speed, making them practical for integration into medical practice.

\subsection{RQ1: Reliability of Current Med-VQA Evaluation}

\paragraph{Adversarial Evaluation in VQA-RAD.}
To assess whether current Med-VQA evaluations capture model vulnerabilities, we first introduce adversarial pairs on the VQA-RAD test set~\citep{lau2018dataset}. Because VQA-RAD provides only finalized QA pairs without detailed metadata, adversarial pairs were manually constructed by medical experts for 118 “yes” instances (yielding 236 QA pairs total). As shown in Table~\ref{table:ablation_adv} (left), despite being based on limited information and scale, these adversarial questions lead to drastic accuracy drops. For example, models such as GPT-4o show a reduction from 69.91\% to 55.08\% (a 14.83\% decrease), highlighting the need for robust evaluation protocols.

\paragraph{Adversarial Evaluation in ProbMed.}
The ProbMed dataset systematically incorporates adversarial pairs on scale for all 57k QA pairs. Here, each diagnostic entity is paired with a ground-truth question and a corresponding adversarial question. Table~\ref{table:ablation_adv} (right) demonstrates a similar significant impact: even the best-performing models experience a minimum 20.00\% drop in accuracy (with an average decrease of 27.78\% across models) when evaluated under this rigorous scheme. This result emphasizes that high accuracy on standard benchmarks can be misleading without essential adversarial evaluation for uncovering model weaknesses.

\begin{table*}[htbp]
\small
\centering
\caption{Categorical and overall accuracy (\%) of different models aggregated among all image types in ProbMed (averaging over three runs). 
The overall accuracy is weighted by the number of questions in each type. 
The best result in each question category is \textbf{in-bold}, and the second best is \underline{underlined}.}
\resizebox{0.9\linewidth}{!}{
\begin{tabular}{@{}lrrrrrr@{}} 
\toprule
\multirow{2.5}{*}{Models}
& \multicolumn{2}{c}{General Question} & \multicolumn{3}{c}{Specialized Question}     & \multirow{2.5}{*}{Overall} \\ 
\cmidrule(r){2-3}
\cmidrule(r){4-6}             
                  & Modality          & Organ            & Abnormality    & Condition/Finding & Position         \\ 
\midrule
\textcolor{gray}{Random Choice} & \textcolor{gray}{25.00}   & \textcolor{gray}{25.00}        & \textcolor{gray}{50.00}          & \textcolor{gray}{35.67}      & \textcolor{gray}{36.48}    & \textcolor{gray}{32.13} \\
\midrule
LLaVA-v1          & 25.30$_{\pm1.18}$ & 41.92$_{\pm1.21}$ & 50.00$_{\pm2.01}$ & 0.35$_{\pm0.03}$ & 0.14$_{\pm0.06}$ & 19.30$_{\pm0.18}$          \\
LLaVA-v1.6        & 6.95$_{\pm0.24}$  & \textbf{80.33}$_{\pm0.34}$ & 45.89$_{\pm0.24}$ & 3.67$_{\pm0.10}$ & 1.37$_{\pm0.17}$ & 24.96$_{\pm0.11}$          \\
MiniGPT-v2        & 3.25$_{\pm0.13}$ & \underline{76.95}$_{\pm0.59}$ & 50.08$_{\pm0.84}$ & 15.23$_{\pm0.76}$ & 7.96$_{\pm0.79}$ & 27.67$_{\pm0.25}$          \\
\midrule
LLaVA-Med-v1      & 5.72$_{\pm0.21}$ & 34.36$_{\pm1.21}$ & 38.30$_{\pm2.83}$ & 20.79$_{\pm0.47}$ & 5.22$_{\pm1.10}$ & 17.90$_{\pm0.38}$          \\
LLaVA-Med-v1.5    & 56.14$_{\pm0.90}$ & 67.96$_{\pm0.08}$ & 49.12$_{\pm0.05}$ & 21.91$_{\pm0.06}$ & 11.65$_{\pm0.03}$ & 40.19$_{\pm0.13}$          \\
CheXagent         & 37.25$_{\pm0.50}$ & 33.75$_{\pm0.17}$ & \textbf{73.31}$_{\pm0.01}$ & 28.52$_{\pm0.08}$ & 7.48$_{\pm0.06}$  & 30.61$_{\pm0.02}$          \\
BiomedGPT         & 60.25$_{\pm0.27}$ & 46.81$_{\pm0.62}$ & 50.31$_{\pm0.24}$ & 14.13$_{\pm0.90}$ & 6.11$_{\pm0.23}$ & 33.34$_{\pm0.17}$          \\
Med-Flamingo      & 44.38$_{\pm0.20}$ & 62.02$_{\pm0.54}$ & 50.00$_{\pm0.01}$ & 26.17$_{\pm0.13}$ & 5.72$_{\pm0.06}$ & 35.66$_{\pm0.14}$          \\
RadFM             & 83.72$_{\pm0.26}$ & 41.04$_{\pm0.33}$ & 60.83$_{\pm0.32}$ & 23.05$_{\pm0.14}$ & 9.10$_{\pm0.29}$ & 41.00$_{\pm0.19}$           \\
\midrule
Gemini Pro        & \underline{96.47}$_{\pm0.88}$ & 75.69$_{\pm1.89}$ & \underline{60.29}$_{\pm1.99}$ & 27.93$_{\pm1.82}$ & 18.44$_{\pm0.77}$ & 55.08$_{\pm0.93}$    \\ 
GPT-4V            & 92.51$_{\pm1.10}$ & 71.73$_{\pm2.45}$ & 53.30$_{\pm1.90}$ & \textbf{35.19}$_{\pm1.16}$ & \underline{22.40}$_{\pm1.89}$ & \underline{55.28}$_{\pm0.98}$ \\
GPT-4o            & \textbf{97.03}$_{\pm0.34}$ & 68.13$_{\pm1.15}$ & 61.79$_{\pm2.28}$ & \underline{29.30}$_{\pm2.55}$ & \textbf{24.06}$_{\pm1.80}$    & \textbf{55.60}$_{\pm1.05}$ \\
\bottomrule
\end{tabular}
}
\label{table:aggre_accuracy}
\end{table*}

\subsection{How Reliable Are LMMs in Medical Diagnosis?}

After "correcting" inflated model accuracy by introducing adversarial pairs, we continue to address the second research question. We conducted diagnostic probing ranging from general to specialized diagnostic questions using the ProbMed dataset.

\paragraph{Performance across Diagnostic Questions}
Table~\ref{table:aggre_accuracy} shows the categorical accuracy of different models aggregated among all image types. 
While GPT-4o, GPT-4V, and Gemini Pro outperform other models and excel in general tasks such as recognizing image modality and organs, their low performance in specialized tasks like determining the existence of abnormalities and answering fine-grained questions about condition/finding and position highlights a significant gap in their ability to aid in real-life diagnosis.

On more specialized diagnostic questions, even top-performing models like GPT-4o, GPT-4V, and Gemini Pro performed close to random guessing. Their accuracy in identifying conditions and positions was alarmingly low, underscoring their limitations in handling fine-grained medical inquiries. 
RadFM, LLaVA-Med-v1.5, and Med-Flamingo outperform other specialized models in general questions, yet still struggle with specialized questions. 
LLaVA-Med-v1.5 achieves much higher accuracy among open-sourced models in identifying conditions/finding and their positions, but still performs around 10\% worse than the proprietary models.

Among the open-sourced general-purpose models, MiniGPT-v2 performs the best, surpassing domain-specific models such as LLaVA-Med-v1 in determining positions of condition/finding without domain-specific training. 
A more detailed breakdown of the performance of different models on different image types across each question type is available in Appendix~\ref{appendix:breakdown_results}. 
Distribution plots of ground-truth answers and model responses within each question category are available in Appendix~\ref{appendix:distribution}.

\paragraph{Error Analysis in Procedural Diagnosis}

An error analysis was conducted on two top-performing models (GPT-4V and Gemini Pro) across three specialized diagnostic question types: Abnormality, Condition/Finding, and Position. 
The procedural diagnosis is conducted by conditioning performance on later and more challenging questions on models' success on simple prior ones. 
For example, model accuracy under the Abnormality category is calculated within those test samples where the model correctly identifies both imaging modality and organ. And the accuracy under the Condition/Finding category is further conditioned on those success samples from prior categories, including Abnormality.

As shown in Table~\ref{table:error_analysis}, both models show vulnerabilities to hallucination errors, particularly in the later stages of the diagnostic procedure. 
For example, errors under the Abnormality arise either from incorrect responses or a tendency to over-reject challenging questions.
In the Condition/Finding and Position categories, errors are largely due to the acceptance of hallucinated entities. Notably, Gemini Pro is more prone to accepting false conditions and positions, which significantly lowers its strict accuracy in these areas.

\begin{table}[ht]
\small
\centering
\caption{Error Analysis of GPT-4V and Gemini Pro on ProbMed. The table shows the accuracy and types of errors for three specialized question types: Abnormality, Condition/Finding, and Position. Errors are categorized into wrong answers, rejection to answer, denying ground truth, and accepting hallucinations, providing a detailed breakdown of model performance and failure modes.}
\resizebox{\linewidth}{!}{
\begin{tabular}{@{}ccrr@{}}
\toprule
\multirow{2.5}{*}{Question Type}     & \multirow{2.5}{*}{Accuracy and Error Type}      &  \multicolumn{2}{c}{Models}       \\ \cmidrule(l){3-4} 
&               & GPT-4V & Gemini Pro \\ \midrule
\multirow{3}{*}{Abnormality}       & Accuracy      & 66.06  & \textbf{67.05}      \\
& E\_wrong\_answer & 67.47  & 100.00        \\
& E\_reject\_to\_answer     & 32.52  & 0.00          \\ \midrule
\multirow{4}{*}{Condition/Finding} & Accuracy      & 36.90   & \textbf{39.97}      \\
& E\_deny\_ground-truth   & 51.69  & 39.04      \\
& E\_accept\_hallucination & 42.12  & 59.69      \\
& E\_reject\_to\_answer     & 6.18   & 1.26       \\ \midrule
\multirow{4}{*}{Position}          & Accuracy      & \textbf{39.97}  & 26.40       \\
& E\_deny\_ground-truth   & 39.04  & 23.31      \\
& E\_accept\_hallucination & 59.69  & 76.68      \\
& E\_reject\_to\_answer     & 1.26   & 0.00          \\ 
\bottomrule
\end{tabular}
}
\label{table:error_analysis}
\end{table}

\begin{figure*}[ht]
    \centering
    \vspace{-10pt}
    \includegraphics[width=\textwidth]{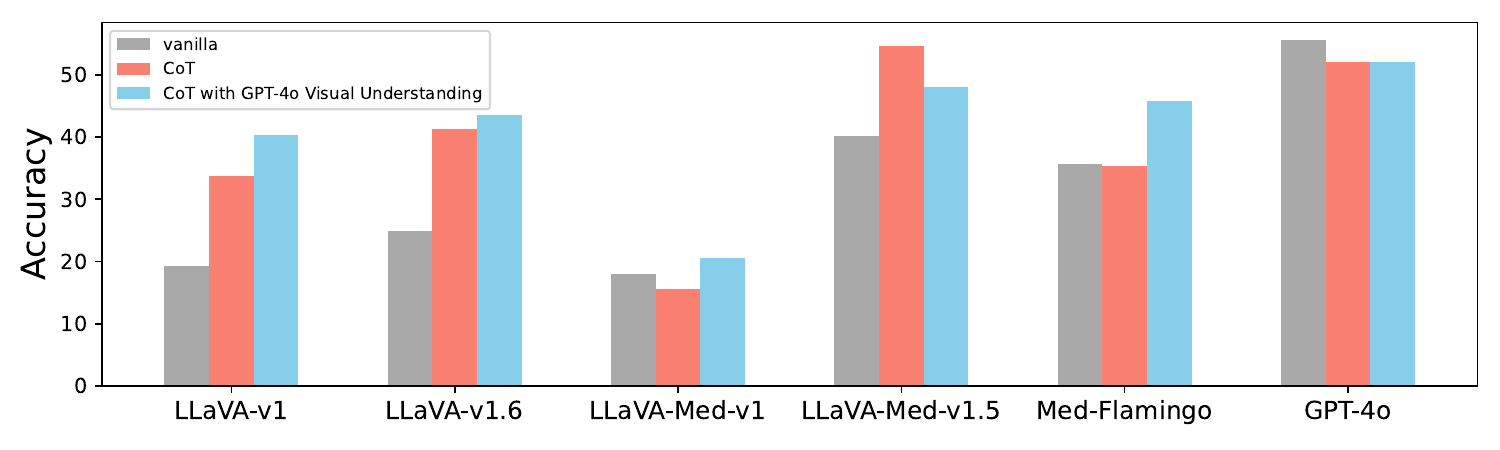}
    \vspace{-15pt}
    \caption{Accuracy comparison of LLaVA-v1, LLaVA-v1.6, LLaVA-Med-v1, LLaVA-Med-v1.5, Med-Flamingo, and GPT-4o under three different settings: vanilla (baseline performance), chain-of-thought (CoT) reasoning, and CoT with GPT-4o-generated visual descriptions. All models demonstrate significant performance improvement when visual descriptions from GPT-4o are included, indicating that poor visual understanding is a key factor limiting baseline performance. Chain-of-thought reasoning alone also leads to notable gains in accuracy, particularly in general-purpose models.}
    \label{fig:ablation}
\end{figure*}

\subsection{Exploring Model Limitations and Potential Improvements}

\paragraph{Impact of Chain-of-Thought Prompting and Visual Understanding}
To further investigate the underperformance of open-source models, we conducted an extensive ablation study on LLaVA-v1, LLaVA-v1.6, LLaVA-Med-v1, LLaVA-Med-v1.5, Med-Flamingo, and GPT-4o. In this study, we examined two additional experimental settings: 
(1) applying a Chain-of-Thought (CoT) approach where models first generate visual descriptions from the image on their own, then utilize the description to augment the question,
(2) enhancing the models by providing external visual descriptions generated by GPT-4o in addition to the question.

As shown in Figure~\ref{fig:ablation}, employing the chain-of-thought approach alone---without external visual descriptions---resulted in an average accuracy increase of 6.51\%. In particular, LLaVA-Med-v1.5's accuracy improved from 40.19\% to 54.55\%, closing the gap to within 1.05\% of the vanilla GPT-4o model. 
Interestingly, GPT-4o's performance decreased by 3.55\% when the CoT mechanism was applied, potentially indicating that the model already internally employs its own chain-of-thought process.

Notably, all open-source models exhibited improved performance when augmented with visual descriptions generated by GPT-4o, suggesting that their baseline limitations stem primarily from poor visual comprehension. 
On average, these models showed an accuracy improvement of 9.44\% across all question categories. This observation suggests that poor visual understanding is a major limitation of existing models, and augmenting them with external visual reasoning can lead to notable gains. Detailed performance changes of each model, organized by question category, can be found in Appendix~\ref{appendix:ablation_breakdown_results}.

\paragraph{Transferability of Domain Expertise}

\begin{figure}[h!]
    \centering
    \includegraphics[width=0.8\columnwidth]{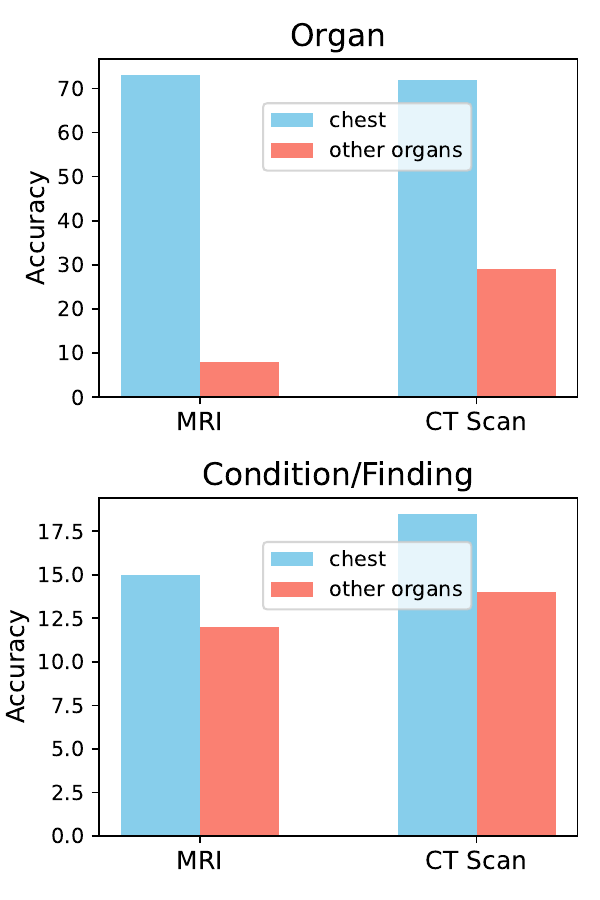}
    \caption{Accuracy comparison of CheXagent in identifying organs and conditions/findings across different modalities. The model demonstrates significantly higher accuracy in identifying organs on chest images compared to images of other organs for both MRI and CT scans. Additionally, CheXagent shows improved accuracy in identifying conditions/findings on chest images, indicating the transferability of its specialized knowledge from chest X-ray training to other imaging modalities.}
    \label{fig:chest_accuracy_transfer}
\end{figure}

As shown in Table~\ref{table:chestxray}, CheXagent, a model trained exclusively on chest X-ray images, performs best in detecting abnormalities and identifying conditions/findings among all models when tested on chest X-ray images. We conducted a finer-grained analysis to explore whether the model's expertise in identifying features of a particular organ can be transferred to other imaging modalities.

As illustrated in Figure~\ref{fig:chest_accuracy_transfer}, CheXagent achieves significantly higher accuracy in identifying chest-related features compared to other organs, confirming our assumption that the model's pre-training on chest X-rays enhances its performance on recognizing chest images across different modalities. Interestingly, CheXagent also demonstrated higher accuracy in identifying conditions and findings in CT scans and MRIs of the chest, achieving a 3\% increase in accuracy for MRIs and a 4\% increase for CT scans compared with other organs within the same unseen modality. This indicates that specialized knowledge gained on chest X-rays can be transferred to other imaging modalities of the same organ in a zero-shot manner, highlighting the potential for cross-modality expertise transfer in real-life medical imaging diagnostics.

\section{Conclusion}

Evaluating the reliability of LMMs in the medical domain requires robust methods, and ProbMed, our newly introduced framework, addresses this by incorporating probing evaluation and procedural diagnosis. 
Our study reveals significant limitations in models like GPT-4o and Gemini Pro, which perform worse than random guessing on specialized diagnostic questions, while CheXagent's results highlight the critical importance of domain-specific knowledge. 
Furthermore, our additional experiments, which introduced Chain-of-Thought reasoning with external visual descriptions generated by GPT-4o, suggested that poor visual understanding is a major limitation of existing models and augmenting them with external visual reasoning can lead to notable gains.

\section{Limitations}
Despite the contributions, limitations such as the imbalanced image distribution favoring Chest X-rays (see Table ~\ref{table:statistics}) and the absence of open-ended evaluations, such as report generation, remain. 
The broader impact of our work includes the potential for improved diagnostic accuracy and better patient care, but it also highlights the risks of deploying unreliable models in healthcare. 
We recommend rigorous testing, continuous performance monitoring, and the incorporation of domain-specific expertise to mitigate these risks. Ultimately, our work aims to contribute to the development of trustworthy AI systems in healthcare, advancing diagnostic outcomes and patient safety.

\section*{Acknowledgments}

This research project has benefited from the Microsoft Accelerate Foundation Models Research (AFMR) grant program.

\bibliography{main}

\newpage
\appendix
\section{Breakdown Results on Different Image Modality and Organ.}
\label{appendix:breakdown_results}
This section shows a detailed model performance breakdown by each (imaging modality, organ) pair under five categories.

\subsection{Brain CT Scan}
\begin{table}[H]
\centering
\caption{Results of different models on Brain CT scan in ProbMed. The best-performing model in each question category is \textbf{in-bold}, and the second best is \underline{underlined}.}
\resizebox{\linewidth}{!}{
\begin{tabular}{@{}ccccccc@{}}
\toprule
& & \multicolumn{2}{c}{General Question} & \multicolumn{3}{c}{Specialized Question} \\
\cmidrule(r){3-4}
\cmidrule(r){5-7}
& & Modality & Organ & Abnormality & Condition/Finding & Position \\ 
\midrule
\multirow{1}{*}{\textcolor{gray}{Random Choice}}   
& \textcolor{gray}{Acc. with adv. pairs} & \textcolor{gray}{25} & \textcolor{gray}{25} & \textcolor{gray}{50} & \textcolor{gray}{35.28} & \textcolor{gray}{35.01}\\
\midrule 
\multirow{2}{*}{LLaVA-v1}   & Acc. with adv. pairs               
                               &25.18   &52.59  &50    &0       &0\\
                               & Avg. acc.
                               &62.59	 &72.22	 &/	    &46.57   &49.60\\
\multirow{2}{*}{LLaVA-v1.6} & Acc. with adv. pairs               
                               &10.74	 &72.22	 &23.52	&0       &0.52\\
                               & Avg. acc.
                               &55.37	 &84.44	 &/	    &30.79   &41.91\\
\multirow{2}{*}{MiniGPT-v2}    & Acc. with adv. pairs               
                               &1.11	 &\textbf{92.59}	 &50	&17.77   &8.42\\
                               & Avg. acc.
                               &50.55	 &96.29	 &/	    &51.20   &54.25\\
\midrule                               
\multirow{2}{*}{LLaVA-Med-v1}  & Acc. with adv. pairs               
                               &4.81	 &10.74	 &8.82	&11.85   &3.15\\
                               & Avg. acc.
                               &50.18	 &33.88	 &/	    &40.71   &49.78\\
\multirow{2}{*}{LLaVA-Med-v1.5}& Acc. with adv. pairs               
                               &50.37	 &80.37	 &44.11	&11.85   &15.26\\
                               & Avg. acc.
                               &74.81	 &89.62	 &/	    &52.98   &54.83\\
\multirow{2}{*}{BiomedGPT}  & Acc. with adv. pairs               
                               &24.44	 &5.18	 &58.82 &14.44   &2.63\\
                               & Avg. acc.
                               &62.03	 &52.59	 &/	    &53.88   &35.84\\   
\multirow{2}{*}{Med-Flamingo}  & Acc. with adv. pairs               
                               &3.70	 &9.62	 &50 &18.14   &5.26\\
                               & Avg. acc.
                               &51.85	 &47.03	 &/	    &50.16   &47.85\\
\multirow{2}{*}{CheXagent}     & Acc. with adv. pairs               
                               &11.85	 &0	    &47.05	&12.96   &5.26\\
                               & Avg. acc.
                               &40.55	 &23.88	 &/	    &53.00   &51.46\\
\midrule                       
\multirow{2}{*}{GPT-4o}        & Acc. with adv. pairs               
                               &\textbf{94.81}	 &\textbf{93.70}	&\underline{61.76}	&\underline{35.92}   &\underline{26.31}\\
                               & Avg. acc.
                               &97.22	 &96.66	&/	    &68.76   &64.83\\
\multirow{2}{*}{GPT-4V}        & Acc. with adv. pairs               
                               &\underline{94.07}	 &84.07	&\underline{61.76}	&\textbf{37.03}   &\textbf{31.05}\\
                               & Avg. acc.
                               &96.85	 &91.48	&/	    &67.01   &65.00\\
\multirow{2}{*}{Gemini Pro}    & Acc. with adv. pairs               
                               &84.44	 &\underline{85.18}	&\textbf{70.58}	&34.81  &21.05\\
                               & Avg. acc.
                               &92.03	 &92.40	&/	    &68.01   &60.16\\
                               \cmidrule(l){2-7} 
\multicolumn{1}{l}{}           & num    
                               & 270    & 270  & 34    &270     &270\\ \bottomrule
\end{tabular}
}
\end{table}

\subsection{Chest CT Scan}
\begin{table}[H]
\centering
\caption{Results of different models on Chest CT Scan in ProbMed. The best-performing model in each question category is \textbf{in-bold}, and the second best is \underline{underlined}.}
\resizebox{\linewidth}{!}{
\begin{tabular}{@{}ccccccc@{}}
\toprule
& & \multicolumn{2}{c}{General Question} & \multicolumn{3}{c}{Specialized Question} \\
\cmidrule(r){3-4}
\cmidrule(r){5-7}
& & Modality & Organ & Abnormality & Condition/Finding & Position \\ 
\midrule
\multirow{1}{*}{\textcolor{gray}{Random Choice}}   
& \textcolor{gray}{Acc. with adv. pairs} & \textcolor{gray}{25} & \textcolor{gray}{25} & \textcolor{gray}{50} & \textcolor{gray}{32.69} & \textcolor{gray}{33.76}\\
\midrule 
\multirow{2}{*}{LLaVA-v1}   & Acc. with adv. pairs               
                               &27.55   &46.35  &50    &0.36    &0.23\\
                               & Avg. acc.
                               &63.77	 &73.08	 &/	    &48.54   &50.11\\
\multirow{2}{*}{LLaVA-v1.6} & Acc. with adv. pairs               
                               &2.73	 &\textbf{76.82}	 &50	&0.54    &0.46\\
                               & Avg. acc.
                               &51.18	 &86.58	 &/	    &41.42   &45.75\\
\multirow{2}{*}{MiniGPT-v2}    & Acc. with adv. pairs               
                               &0.54	 &53.28	 &50	&10.21   &3.22\\
                               & Avg. acc.
                               &50.27	 &75.82	 &/	    &51.11   &51.49\\
\midrule                               
\multirow{2}{*}{LLaVA-Med-v1}  & Acc. with adv. pairs               
                               &5.47	 &39.78	 &29.41	&14.41   &4.37\\
                               & Avg. acc.
                               &51.18	 &68.06	 &/	    &45.50   &51.72\\
\multirow{2}{*}{LLaVA-Med-v1.5}& Acc. with adv. pairs               
                               &51.09	 &61.86	 &41.17	&14.78   &9.21\\
                               & Avg. acc.
                               &75.54	 &80.10	 &/	    &52.60   &54.64\\
\multirow{2}{*}{BiomedGPT}  & Acc. with adv. pairs               
                               &15.51	 &2.91	 &52.94 &7.11    &2.30\\
                               & Avg. acc.
                               &56.93	 &50.63	 &/	    &50.93   &34.65\\ 
\multirow{2}{*}{Med-Flamingo}  & Acc. with adv. pairs               
                               &22.26	 &70.98	 &50    &19.16   &7.14\\
                               & Avg. acc.
                               &60.31	 &85.49	 &/	    &51.11   &48.89\\
\multirow{2}{*}{CheXagent}     & Acc. with adv. pairs               
                               &6.75	 &\underline{72.99}	 &50   	&18.61   &7.83\\
                               & Avg. acc.
                               &32.93	 &86.49	 &/	    &56.80   &51.55\\
\midrule     
\multirow{2}{*}{GPT-4o}        & Acc. with adv. pairs               
                               &\textbf{97.62}	 &65.99	&\underline{67.64}	&27.60    &\underline{19.58}\\
                               & Avg. acc.
                               &98.72	 &81.90	&/	    &63.54   &61.67\\
\multirow{2}{*}{GPT-4V}        & Acc. with adv. pairs               
                               &\underline{97.07}	 &72.94	&\underline{67.64}	&\underline{32.9}    &\textbf{20.78}\\
                               & Avg. acc.
                               &98.44	 &85.74	&/	    &65.01   &59.54\\
\multirow{2}{*}{Gemini Pro}    & Acc. with adv. pairs               
                               &95.62	 &58.21	&\textbf{82.35}	&\textbf{34.48}   &14.28\\
                               & Avg. acc.
                               &97.71	 &78.37	&/	    &65.62   &56.84\\
                               \cmidrule(l){2-7} 
\multicolumn{1}{l}{}           & num    
                               & 548    & 548  & 34    &548     &548\\ \bottomrule
\end{tabular}
}
\end{table}

\subsection{Spine CT Scan}
\begin{table}[H]
\centering
\caption{Results of different models on Spine CT Scan in ProbMed. The best-performing model in each question category is \textbf{in-bold}, and the second best is \underline{underlined}.}
\resizebox{\linewidth}{!}{
\begin{tabular}{@{}ccccccc@{}}
\toprule
& & \multicolumn{2}{c}{General Question} & \multicolumn{3}{c}{Specialized Question} \\
\cmidrule(r){3-4}
\cmidrule(r){5-7}
& & Modality & Organ & Abnormality & Condition/Finding & Position \\ 
\midrule
\multirow{1}{*}{\textcolor{gray}{Random Choice}}   
& \textcolor{gray}{Acc. with adv. pairs} & \textcolor{gray}{25} & \textcolor{gray}{25} & \textcolor{gray}{50} & \textcolor{gray}{30.85} & \textcolor{gray}{31.06}\\
\midrule 
\multirow{2}{*}{LLaVA-v1}   & Acc. with adv. pairs               
                               &22.98   &44.82  &50    &0       &0\\
                               & Avg. acc.
                               &61.49	 &70.68	 &/	    &49.47   &50.00\\
\multirow{2}{*}{LLaVA-v1.6} & Acc. with adv. pairs               
                               &4.59	 &72.41	 &0	    &0       &1.28\\
                               & Avg. acc.
                               &52.29	 &83.90	 &/	    &37.66   &41.07\\
\multirow{2}{*}{MiniGPT-v2}    & Acc. with adv. pairs               
                               &1.14	 &41.37	 &0	    &12.64   &5.12\\
                               & Avg. acc.
                               &50.57	 &58.62	 &/	    &54.41   &51.21\\
\midrule                               
\multirow{2}{*}{LLaVA-Med-v1}  & Acc. with adv. pairs               
                               &2.29	 &11.49	 &50	&11.49   &6.41\\
                               & Avg. acc.
                               &48.27	 &30.45	 &/	    &46.37   &48.77\\
\multirow{2}{*}{LLaVA-Med-v1.5}& Acc. with adv. pairs               
                               &32.18	 &67.81	 &50.0	&9.19    &14.10\\
                               & Avg. acc.
                               &65.51	 &83.33	 &/	    &55.23   &51.27\\
\multirow{2}{*}{BiomedGPT}  & Acc. with adv. pairs               
                               &28.73	 &8.04	 &0 &6.89   &2.56\\
                               & Avg. acc.
                               &63.79	 &53.44	 &/	    &50.00   &33.27\\                                
\multirow{2}{*}{Med-Flamingo}  & Acc. with adv. pairs               
                               &6.89	 &39.08	 &50 &14.94   &8.97\\
                               & Avg. acc.
                               &53.44	 &68.39	 &/	    &53.92   &52.22\\
\multirow{2}{*}{CheXagent}     & Acc. with adv. pairs               
                               &4.59	 &27.58	 &50   	&10.34   &2.56\\
                               & Avg. acc.
                               &34.48	 &58.04	 &/	    &49.45   &50.20\\
\midrule     
\multirow{2}{*}{GPT-4o}        & Acc. with adv. pairs               
                               &\textbf{87.35}	 &\underline{76.74}	 &0	    &\underline{30.23}    &\underline{20.77}\\
                               & Avg. acc.
                               &93.10	 &88.37	 &/	    &66.01   &60.08\\
\multirow{2}{*}{GPT-4V}        & Acc. with adv. pairs               
                               &81.39	 &69.76	 &0	    &\textbf{33.73}    &\textbf{25.97}\\
                               & Avg. acc.
                               &89.53	 &84.30	 &/	    &65.77   &63.13\\
\multirow{2}{*}{Gemini Pro}    & Acc. with adv. pairs               
                               &\underline{87.2}	 &\textbf{77.9}	 &50	&22.09   &\textbf{25.97}\\
                               & Avg. acc.
                               &92.44	 &88.95	 &/	    &61.64   &64.94\\
                               \cmidrule(l){2-7} 
\multicolumn{1}{l}{}           & num    
                               &86      &86     &2     &86      &86\\ \bottomrule
\end{tabular}
}
\end{table}

\subsection{Abdominal CT Scan}
\begin{table}[H]
\centering
\caption{Results of different models on Abdominal CT Scan in ProbMed. The best-performing model in each question category is \textbf{in-bold}, and the second best is \underline{underlined}.}
\resizebox{\linewidth}{!}{
\begin{tabular}{@{}ccccccc@{}}
\toprule
& & \multicolumn{2}{c}{General Question} & \multicolumn{3}{c}{Specialized Question} \\
\cmidrule(r){3-4}
\cmidrule(r){5-7}
& & Modality & Organ & Abnormality & Condition/Finding & Position \\ 
\midrule
\multirow{1}{*}{\textcolor{gray}{Random Choice}}   
& \textcolor{gray}{Acc. with adv. pairs} & \textcolor{gray}{25} & \textcolor{gray}{25} & \textcolor{gray}{50} & \textcolor{gray}{35.53} & \textcolor{gray}{37.03}\\
\midrule 
\multirow{2}{*}{LLaVA-v1}   & Acc. with adv. pairs               
                               &26.49   &54.19  &50    &0.53    &0\\
                               & Avg. acc.
                               &63.24	 &77.09	 &/	    &47.70   &50.00\\
\multirow{2}{*}{LLaVA-v1.6} & Acc. with adv. pairs               
                               &1.86	 &\textbf{82.82}	 &41.42	&1.06    &0.66\\
                               & Avg. acc.
                               &50.93	 &91.07	 &/	    &38.36   &45.82\\
\multirow{2}{*}{MiniGPT-v2}    & Acc. with adv. pairs               
                               &0	     &37.15	 &48.57	&6.12    &2.14\\
                               & Avg. acc.
                               &50.00	 &66.97	 &/	    &48.49   &50.22\\
\midrule                               
\multirow{2}{*}{LLaVA-Med-v1}  & Acc. with adv. pairs               
                               &5.05	 &45	 &30	&15.44   &5.28\\
                               & Avg. acc.
                               &51.53   &70.90	 &/	    &45.13   &49.24\\
\multirow{2}{*}{LLaVA-Med-v1.5}& Acc. with adv. pairs               
                               &51.93	 &67.64	 &48.57 &11.31   &16.03\\
                               & Avg. acc.
                               &75.96	 &83.42	 &/	    &52.86   &65.61\\
\multirow{2}{*}{BiomedGPT}  & Acc. with adv. pairs               
                               &67.77	 &12.38	 &\underline{57.14} &15.31   &4.62\\
                               & Avg. acc.
                               &83.75	 &55.52	 &/	    &54.49   &45.06\\         
\multirow{2}{*}{Med-Flamingo}  & Acc. with adv. pairs               
                               &1.73	 &35.55	 &50 &20.37   &8.26\\
                               & Avg. acc.
                               &50.86	 &67.57	 &/	    &51.03   &49.46\\               
\multirow{2}{*}{CheXagent}     & Acc. with adv. pairs               
                               &25.03	 &38.21	 &52.85 &15.57   &6.61\\
                               & Avg. acc.
                               &51.46	 &65.71	 &/	    &51.08   &50.19\\
\midrule                               
\multirow{2}{*}{GPT-4o}        & Acc. with adv. pairs               
                               &\underline{97.99}	 &65.28	 &51.42	&23.12      &\textbf{28.23}\\
                               & Avg. acc.
                               &98.93	 &81.50	 &/	    &58.24    &64.59\\                               
\multirow{2}{*}{GPT-4V}        & Acc. with adv. pairs               
                               &95.72	 &\underline{72.72}	 &45.71	&\underline{27}      &\underline{23.25}\\
                               & Avg. acc.
                               &97.72	 &85.56	 &/	    &58.92    &60.02\\
\multirow{2}{*}{Gemini Pro}    & Acc. with adv. pairs               
                               &\textbf{98.31}	 &69.19	 &\textbf{65.71}	&\textbf{28.79}   &20.39\\
                               & Avg. acc.
                               &99.00	 &84.20	 &/	    &61.03    &59.27\\
                               \cmidrule(l){2-7} 
\multicolumn{1}{l}{}           & num    
                               &750     &750    &70    &750     &750\\ \bottomrule
\end{tabular}
}
\end{table}

\subsection{Brain MRI}
\begin{table}[H]
\centering
\caption{Results of different models on Brain MRI in ProbMed. The best-performing model in each question category is \textbf{in-bold}, and the second best is \underline{underlined}.}
\resizebox{\linewidth}{!}{
\begin{tabular}{@{}ccccccc@{}}
\toprule
& & \multicolumn{2}{c}{General Question} & \multicolumn{3}{c}{Specialized Question} \\
\cmidrule(r){3-4}
\cmidrule(r){5-7}
& & Modality & Organ & Abnormality & Condition/Finding & Position \\ 
\midrule
\multirow{1}{*}{\textcolor{gray}{Random Choice}}   
& \textcolor{gray}{Acc. with adv. pairs} & \textcolor{gray}{25} & \textcolor{gray}{25} & \textcolor{gray}{50} & \textcolor{gray}{36.7} & \textcolor{gray}{36.64}\\
\midrule 
\multirow{2}{*}{LLaVA-v1}   & Acc. with adv. pairs               
                               &1.23    &32.86  &50    &0.53    &0\\
                               & Avg. acc.
                               &49.29	 &65.37	 &/	    &47.85   &49.87\\
\multirow{2}{*}{LLaVA-v1.6} & Acc. with adv. pairs               
                               &17.49	 &88.51	 &28.57	&0.53    &0.48\\
                               & Avg. acc.
                               &58.74	 &93.10	 &/	    &31.73   &37.46\\
\multirow{2}{*}{MiniGPT-v2}    & Acc. with adv. pairs               
                               &1.94	 &\underline{96.64}	 &50	&15.72    &4.37\\
                               & Avg. acc.
                               &50.88	 &98.32	 &/	    &52.16   &50.51\\
\midrule                               
\multirow{2}{*}{LLaVA-Med-v1}  & Acc. with adv. pairs               
                               &3	     &8.12	 &23.21	&14.66   &2.91\\
                               & Avg. acc.
                               &47.08	 &32.50	 &/	    &47.72   &48.35\\
\multirow{2}{*}{LLaVA-Med-v1.5}& Acc. with adv. pairs               
                               &75.61	 &84.98	 &42.85	&13.78   &13.62\\
                               & Avg. acc.
                               &87.80	 &92.40	 &/	    &53.37   &53.52\\
\multirow{2}{*}{BiomedGPT}  & Acc. with adv. pairs               
                               &15.37	 &12.36	 &44.64 &11.48   &2.67\\
                               & Avg. acc.
                               &54.41	 &56.00	 &/	    &51.26   &42.06\\ 
\multirow{2}{*}{Med-Flamingo}  & Acc. with adv. pairs               
                               &0.35	 &13.60	 &50 &10.77   &3.16\\
                               & Avg. acc.
                               &47.61	 &51.32	 &/	    &48.27   &50.01\\
\multirow{2}{*}{CheXagent}     & Acc. with adv. pairs               
                               &0	     &0	     &50    &10.77   &6.81\\
                               & Avg. acc.
                               &20.40	 &21.99	 &/	    &50.37   &51.87\\
\midrule     
\multirow{2}{*}{GPT-4o}        & Acc. with adv. pairs               
                               &\textbf{97.69}	 &\textbf{97.34}	 &\underline{66.07}	&25.84   &\textbf{30.24}\\
                               & Avg. acc.
                               &98.58	 &98.67	 &/	    &61.05   &66.13\\
\multirow{2}{*}{GPT-4V}        & Acc. with adv. pairs               
                               &\underline{96.99}	 &94.33	 &58.92	&\textbf{36.1}   &\underline{27.8}\\
                               & Avg. acc.
                               &98.40	 &97.07	 &/	    &65.89   &62.38\\
\multirow{2}{*}{Gemini Pro}    & Acc. with adv. pairs               
                               &95.22	 &94.87	 &\textbf{78.57}	&\underline{35.51}   &19.7\\
                               & Avg. acc.
                               &97.26	 &97.34	 &/	    &65.59   &59.67\\
                               \cmidrule(l){2-7} 
\multicolumn{1}{l}{}           & num    
                               &566     &566    &56    &566     &566\\ \bottomrule
\end{tabular}
}
\end{table}

\subsection{Chest MRI}
\begin{table}[H]
\centering
\caption{Results of different models on Chest MRI in ProbMed. The best-performing model in each question category is \textbf{in-bold}, and the second best is \underline{underlined}.}
\resizebox{\linewidth}{!}{
\begin{tabular}{@{}ccccccc@{}}
\toprule
& & \multicolumn{2}{c}{General Question} & \multicolumn{3}{c}{Specialized Question} \\
\cmidrule(r){3-4}
\cmidrule(r){5-7}
& & Modality & Organ & Abnormality & Condition/Finding & Position \\ 
\midrule
\multirow{1}{*}{\textcolor{gray}{Random Choice}}   
& \textcolor{gray}{Acc. with adv. pairs} & \textcolor{gray}{25} & \textcolor{gray}{25} & \textcolor{gray}{50} & \textcolor{gray}{34.18} & \textcolor{gray}{34.11}\\
\midrule 
\multirow{2}{*}{LLaVA-v1}   & Acc. with adv. pairs               
                               &0       &35     &50    &0       &0\\
                               & Avg. acc.
                               &41.25	 &66.25	 &/	    &45.00   &50.00\\
\multirow{2}{*}{LLaVA-v1.6} & Acc. with adv. pairs               
                               &5	     &32.5	 &37.5	&0       &0\\
                               & Avg. acc.
                               &51.24   &56.25	 &/	    &31.35   &43.01\\
\multirow{2}{*}{MiniGPT-v2}    & Acc. with adv. pairs               
                               &0	     &35	 &50	&10      &8.82\\
                               & Avg. acc.
                               &47.50	 &62.50	 &/	    &47.91   &49.50\\
\midrule                               
\multirow{2}{*}{LLaVA-Med-v1}  & Acc. with adv. pairs               
                               &5	     &45	 &12.5	&12.5    &5.88\\
                               & Avg. acc.
                               &43.75	 &68.75	 &/	    &49.06   &46.32\\
\multirow{2}{*}{LLaVA-Med-v1.5}& Acc. with adv. pairs               
                               &50.00   &35.00	 &50.00	&12.5    &11.76\\
                               & Avg. acc.
                               &72.5	 &62.5	 &/	    &53.75   &53.92\\
\multirow{2}{*}{BiomedGPT}  & Acc. with adv. pairs               
                               &0.00	 &5.00	 &50.00 &10.00   &2.94\\
                               & Avg. acc.
                               &40.00	 &51.24	 &/	    &51.04   &49.01\\ 
\multirow{2}{*}{Med-Flamingo}  & Acc. with adv. pairs               
                               &2.50	 &45.00	 &50 &10.00   &8.82\\
                               & Avg. acc.
                               &43.75	 &72.50	 &/	    &48.75   &47.79\\                               
\multirow{2}{*}{CheXagent}     & Acc. with adv. pairs               
                               &0	     &\textbf{75}	 &50    &15      &0\\
                               & Avg. acc.
                               &17.50   &87.50	 &/	    &44.58   &47.05\\
\midrule     
\multirow{2}{*}{GPT-4o}        & Acc. with adv. pairs               
                               &\textbf{90.00}	 &35.89	 &\textbf{62.50}	&\underline{17.94}   &\textbf{24.24}\\
                               & Avg. acc.
                               &93.75	 &65.38	 &/	    &54.80   &61.36\\
\multirow{2}{*}{GPT-4V}        & Acc. with adv. pairs               
                               &76.92	 &\underline{51.28}	 &37.5	&\textbf{25.64}   &\underline{18.18}\\
                               & Avg. acc.
                               &86.25	 &71.79	 &/	    &58.11   &61.74\\
\multirow{2}{*}{Gemini Pro}    & Acc. with adv. pairs               
                               &\underline{87.5}	 &\textbf{62.5}	 &37.5	&17.5    &11.76\\
                               & Avg. acc.
                               &91.25	 &77.50	 &/	    &54.89   &56.86\\
                               \cmidrule(l){2-7} 
\multicolumn{1}{l}{}           & num    
                               &40      &40     &8     &40      &40\\ \bottomrule
\end{tabular}
}
\end{table}

\subsection{Spine MRI}
\begin{table}[H]
\centering
\caption{Results of different models on Spine MRI in ProbMed. The best-performing model in each question category is \textbf{in-bold}, and the second best is \underline{underlined}.}
\resizebox{\linewidth}{!}{
\begin{tabular}{@{}ccccccc@{}}
\toprule
& & \multicolumn{2}{c}{General Question} & \multicolumn{3}{c}{Specialized Question} \\
\cmidrule(r){3-4}
\cmidrule(r){5-7}
& & Modality & Organ & Abnormality & Condition/Finding & Position \\ 
\midrule
\multirow{1}{*}{\textcolor{gray}{Random Choice}}   
& \textcolor{gray}{Acc. with adv. pairs} & \textcolor{gray}{25} & \textcolor{gray}{25} & \textcolor{gray}{50} & \textcolor{gray}{31.51} & \textcolor{gray}{31.52}\\
\midrule 
\multirow{2}{*}{LLaVA-v1}   & Acc. with adv. pairs               
                               &0       &32.09  &50    &50      &0.3\\
                               & Avg. acc.
                               &49.22	 &65.58	 &/	    &47.15   &49.88\\
\multirow{2}{*}{LLaVA-v1.6} & Acc. with adv. pairs               
                               &3.08	 &86.72	 &27.77	&0.30    &0\\
                               & Avg. acc.
                               &51.54   &92.74	 &/	    &30.59   &34.37\\
\multirow{2}{*}{MiniGPT-v2}    & Acc. with adv. pairs               
                               &0.3	 &49.69	 &52.77	&6.79    &2.02\\
                               & Avg. acc.
                               &50.15	 &70.52	 &/	    &48.97   &50.01\\
\midrule                               
\multirow{2}{*}{LLaVA-Med-v1}  & Acc. with adv. pairs               
                               &1.54	 &5.24	 &36.11	&12.96   &5.4\\
                               & Avg. acc.
                               &45.06	 &24.07	 &/	    &48.52   &48.04\\
\multirow{2}{*}{LLaVA-Med-v1.5}& Acc. with adv. pairs               
                               &70.67	 &84.56	 &50.00	&11.11   &11.14\\
                               & Avg. acc.
                               &84.72	 &91.97	 &/	    &52.40   &51.89\\
\multirow{2}{*}{BiomedGPT}  & Acc. with adv. pairs               
                               &0.30	 &5.86	 &50.00 &7.71    &3.04\\
                               & Avg. acc.
                               &45.06	 &52.77	 &/	    &51.31   &44.04\\ 
\multirow{2}{*}{Med-Flamingo}  & Acc. with adv. pairs               
                               &0.30	 &29.93	 &50 `` &17.90   &5.40\\
                               & Avg. acc.
                               &50.00	 &64.50	 &/	    &50.54   &50.14\\                               
\multirow{2}{*}{CheXagent}     & Acc. with adv. pairs               
                               &0	     &13.58	 &47.22 &15.43   &2.7\\
                               & Avg. acc.
                               &22.53	 &44.44	 &/	    &51.28   &48.54\\
\midrule   
\multirow{2}{*}{GPT-4o}        & Acc. with adv. pairs               
                               &\textbf{98.44}	 &84.52	 &\textbf{63.88}	&19.50    &\textbf{24.40}\\
                               & Avg. acc.
                               &98.91	 &91.95	 &/	    &55.46   &63.70\\
\multirow{2}{*}{GPT-4V}        & Acc. with adv. pairs               
                               &96.28	 &\textbf{90.71}	 &55.55	&\underline{22.6}    &\underline{15.59}\\
                               & Avg. acc.
                               &97.51	 &94.73	 &/	    &58.89   &57.52\\
\multirow{2}{*}{Gemini Pro}    & Acc. with adv. pairs               
                               &\underline{98.13}	 &\underline{88.81}	 &\underline{57.14}	&\textbf{24.53}   &14.91\\
                               & Avg. acc.
                               &98.75	 &94.09	 &/	    &59.19   &58.20\\
                               \cmidrule(l){2-7} 
\multicolumn{1}{l}{}           & num    
                               &332     &332    &35    &332     &332\\ \bottomrule
\end{tabular}
}
\end{table}

\subsection{Abdominal MRI}
\begin{table}[H]
\centering
\caption{Results of different models on Abdominal MRI in ProbMed. The best-performing model in each question category is \textbf{in-bold}, and the second best is \underline{underlined}.}
\resizebox{\linewidth}{!}{
\begin{tabular}{@{}ccccccc@{}}
\toprule
& & \multicolumn{2}{c}{General Question} & \multicolumn{3}{c}{Specialized Question} \\
\cmidrule(r){3-4}
\cmidrule(r){5-7}
& & Modality & Organ & Abnormality & Condition/Finding & Position \\ 
\midrule
\multirow{1}{*}{\textcolor{gray}{Random Choice}}   
& \textcolor{gray}{Acc. with adv. pairs} & \textcolor{gray}{25} & \textcolor{gray}{25} & \textcolor{gray}{50} & \textcolor{gray}{37.13} & \textcolor{gray}{38.26}\\
\midrule 
\multirow{2}{*}{LLaVA-v1}   & Acc. with adv. pairs               
                               &0       &39.28  &50.00 &2.38    &0\\
                               & Avg. acc.
                               &48.22	 &69.64	 &/	    &46.42   &50.00\\
\multirow{2}{*}{LLaVA-v1.6} & Acc. with adv. pairs               
                               &2.38	 &\underline{73.8}	 &35.71	&1.19    &0\\
                               & Avg. acc.
                               &51.19   &85.11	 &/	    &35.46   &44.06\\
\multirow{2}{*}{MiniGPT-v2}    & Acc. with adv. pairs               
                               &0  	 &36.9	 &50	&8.33    &4.54\\
                               & Avg. acc.
                               &50.00	 &67.26	 &/	    &47.51   &51.70\\
\midrule                               
\multirow{2}{*}{LLaVA-Med-v1}  & Acc. with adv. pairs               
                               &2.38	 &47.61	 &50.00	&14.28   &9.09\\
                               & Avg. acc.
                               &41.66	 &72.61	 &/	    &47.42   &46.46\\
\multirow{2}{*}{LLaVA-Med-v1.5}& Acc. with adv. pairs               
                               &51.19	 &65.47	 &50.00	&13.09   &16.66\\
                               & Avg. acc.
                               &75.59	 &81.54	 &/	    &54.31   &56.37\\
\multirow{2}{*}{BiomedGPT}  & Acc. with adv. pairs               
                               &1.19	 &3.57	 &50.00 &14.28   &1.51\\
                               & Avg. acc.
                               &38.69	 &50.00	 &/	    &51.33   &46.46\\                                
\multirow{2}{*}{Med-Flamingo}  & Acc. with adv. pairs               
                               &2.38	 &27.38	 &50.00 &20.23   &3.03\\
                               & Avg. acc.
                               &50.59	 &62.50	 &/	    &49.55   &50.50\\
\multirow{2}{*}{CheXagent}     & Acc. with adv. pairs               
                               &0	     &26.19	 &50.00 &11.9    &10.6\\
                               & Avg. acc.
                               &19.04   &56.54	 &/	    &49.20   &49.62\\
\midrule  
\multirow{2}{*}{GPT-4o}        & Acc. with adv. pairs               
                               &\textbf{91.66}	 &67.85	 &\underline{64.28}	&21.42 &\textbf{39.39}\\
                               & Avg. acc.
                               &95.83	 &81.54	 &/	    &55.30    &70.51\\
\multirow{2}{*}{GPT-4V}        & Acc. with adv. pairs               
                               &86.9	 &\textbf{75}	 &50	&\underline{27.38}   &\underline{25.75}\\
                               & Avg. acc.
                               &92.26	 &85.71	 &/	    &58.58     &58.77\\
\multirow{2}{*}{Gemini Pro}    & Acc. with adv. pairs               
                               &\underline{89.28}	 &72.61	 &\textbf{85.71}	&\textbf{28.57}   &\underline{25.75}\\
                               & Avg. acc.
                               &94.04	 &86.30	 &/	    &63.39   &60.98\\
                               \cmidrule(l){2-7} 
\multicolumn{1}{l}{}           & num    
                               &84      &84     &14    &84      &84\\ \bottomrule
\end{tabular}
}
\end{table}

\subsection{Brain X-ray}
\begin{table}[H]
\centering
\caption{Results of different models on Brain X-ray in ProbMed. The best-performing model in each question category is \textbf{in-bold}, and the second best is \underline{underlined}.}
\resizebox{\linewidth}{!}{
\begin{tabular}{@{}ccccccc@{}}
\toprule
& & \multicolumn{2}{c}{General Question} & \multicolumn{3}{c}{Specialized Question} \\
\cmidrule(r){3-4}
\cmidrule(r){5-7}
& & Modality & Organ & Abnormality & Condition/Finding & Position \\ 
\midrule
\multirow{1}{*}{\textcolor{gray}{Random Choice}}   
& \textcolor{gray}{Acc. with adv. pairs} & \textcolor{gray}{25} & \textcolor{gray}{25} & \textcolor{gray}{50} & \textcolor{gray}{44.77} & \textcolor{gray}{47.08}\\
\midrule 
\multirow{2}{*}{LLaVA-v1}   & Acc. with adv. pairs               
                               &45.56   &26.58  &50    &0       &0\\
                               & Avg. acc.
                               &72.78	 &51.89	 &/	    &48.10   &50.00\\
\multirow{2}{*}{LLaVA-v1.6} & Acc. with adv. pairs               
                               &11.39	 &13.92	 &16.66	&8.86    &4.44\\
                               & Avg. acc.
                               &55.06   &48.10	 &/	    &45.04   &48.88\\
\multirow{2}{*}{MiniGPT-v2}    & Acc. with adv. pairs               
                               &18.98   &\textbf{83.54} &50	&18.98   &17.77\\
                               & Avg. acc.
                               &59.49	 &89.87	 &/	    &51.37   &52.22\\
\midrule                               
\multirow{2}{*}{LLaVA-Med-v1}  & Acc. with adv. pairs               
                               &8.86	 &8.86	 &0	    &20.25   &4.44\\
                               & Avg. acc.
                               &54.43	 &31.01	 &/	    &51.16   &48.33\\
\multirow{2}{*}{LLaVA-Med-v1.5}& Acc. with adv. pairs               
                               &49.36	 &31.64	 &50.00	&8.86    &13.33\\
                               & Avg. acc.
                               &73.41	 &56.96	 &/	    &53.16   &55.55\\
\multirow{2}{*}{BiomedGPT}  & Acc. with adv. pairs               
                               &12.65	 &6.32	 &50.00 &11.39   &2.22\\
                               & Avg. acc.
                               &53.16	 &49.36	 &/	    &52.95   &43.33\\                                
\multirow{2}{*}{Med-Flamingo}  & Acc. with adv. pairs               
                               &8.86	 &0	 &50 &22.78   &8.88\\
                               & Avg. acc.
                               &54.43	 &15.18	 &/	    &50.73   &48.33\\                               
\multirow{2}{*}{CheXagent}     & Acc. with adv. pairs               
                               &84.81	 &0	     &50    &12.65   &8.88\\
                               & Avg. acc.
                               &92.40   &29.74	 &/	    &51.16   &55.00\\
\midrule  
\multirow{2}{*}{GPT-4o}        & Acc. with adv. pairs               
                               &\textbf{94.93}	 &\underline{52.56}	 &\textbf{66.66}	&\underline{37.17}   &\textbf{40.90}\\
                               & Avg. acc.
                               &96.20	 &73.71	 &/	    &62.07    &69.31\\
\multirow{2}{*}{GPT-4V}        & Acc. with adv. pairs               
                               &82.05	 &8.97	 &33.33	&\textbf{43.58}   &22.72\\
                               & Avg. acc.
                               &90.38	 &47.43	 &/	    &68.48      &59.09\\
\multirow{2}{*}{Gemini Pro}    & Acc. with adv. pairs               
                               &\underline{89.87}	 &51.89	 &50	&31.64   &\underline{31.11}\\
                               & Avg. acc.
                               &93.03	 &74.05	 &/	    &61.81   &63.88\\
                               \cmidrule(l){2-7} 
\multicolumn{1}{l}{}           & num    
                               &79      &79     &6     &79      &79\\ \bottomrule
\end{tabular}
}
\end{table}

\subsection{Chest X-ray}
\begin{table}[H]
\centering
\caption{Results of different models on Chest X-ray in ProbMed. The best-performing model in each question category is \textbf{in-bold}, and the second best is \underline{underlined}.}
\resizebox{\linewidth}{!}{
\begin{tabular}{@{}ccccccc@{}}
\toprule
& & \multicolumn{2}{c}{General Question} & \multicolumn{3}{c}{Specialized Question} \\
\cmidrule(r){3-4}
\cmidrule(r){5-7}
& & Modality & Organ & Abnormality & Condition/Finding & Position \\ 
\midrule
\multirow{1}{*}{\textcolor{gray}{Random Choice}}   
& \textcolor{gray}{Acc. with adv. pairs} & \textcolor{gray}{25} & \textcolor{gray}{25} & \textcolor{gray}{50} & \textcolor{gray}{37.59} & \textcolor{gray}{37.08}\\
\midrule 
\multirow{2}{*}{LLaVA-v1}   & Acc. with adv. pairs               
                               &28.75   &36.57  &50    &0.12    &0.11\\
                               & Avg. acc.
                               &64.37	 &68.25	 &/	    &34.41   &50.05\\
\multirow{2}{*}{LLaVA-v1.6} & Acc. with adv. pairs               
                               &7.11	 &83.97	 &47.94	&5.89    &1.52\\
                               & Avg. acc.
                               &53.49   &91.61	 &/	    &34.52   &48.85\\
\multirow{2}{*}{MiniGPT-v2}    & Acc. with adv. pairs               
                               &4.93    &\textbf{94.07}	 &50.05	&18.78   &11.94\\
                               & Avg. acc.
                               &52.46	 &96.98	 &/	    &46.09   &53.15\\
\midrule                               
\multirow{2}{*}{LLaVA-Med-v1}  & Acc. with adv. pairs               
                               &6.25	 &39.77	 &40.24	&26.28   &6.14\\
                               & Avg. acc.
                               &52.62 	 &67.19	 &/	    &50.78   &51.34\\
\multirow{2}{*}{LLaVA-Med-v1.5}& Acc. with adv. pairs               
                               &55.44	 &65.48	 &49.53	&31.82   &9.78\\
                               & Avg. acc.
                               &77.67 	 &82.69	 &/	    &62.70   &54.22\\
\multirow{2}{*}{BiomedGPT}  & Acc. with adv. pairs               
                               &91.34	 &86.05	 &50.00 &16.92   &9.08\\
                               & Avg. acc.
                               &95.46	 &92.93	 &/	    &43.00   &41.46\\ 
\multirow{2}{*}{Med-Flamingo}  & Acc. with adv. pairs               
                               &80.92	 &\underline{90.00}	 &50 &35.83   &5.24\\
                               & Avg. acc.
                               &90.46	 &95.00	 &/	    &63.47   &48.00\\
\multirow{2}{*}{CheXagent}     & Acc. with adv. pairs               
                               &53.68	 &39.64	 &\textbf{76.59} &\textbf{42.75}   &9.38\\
                               & Avg. acc.
                               &76.84   &69.82	 &/	    &70.80   &54.00\\
\midrule 
\multirow{2}{*}{GPT-4o}        & Acc. with adv. pairs               
                               &\underline{97.97}	 &62.98	 &\underline{62.01}	&32.13   &\textbf{21.81}\\
                               & Avg. acc.
                               &98.81	 &81.39	 &/	    &59.35   &59.95\\
\multirow{2}{*}{GPT-4V}        & Acc. with adv. pairs               
                               &91.53	 &67.51	 &53.18	&\underline{39.35}   &\underline{21.35}\\
                               & Avg. acc.
                               &95.62	 &83.37	 &/	    &64.69   &55.64\\
\multirow{2}{*}{Gemini Pro}    & Acc. with adv. pairs               
                               &\textbf{98.07}	 &76.74	 &61.29	&25.83   &15.31\\
                               & Avg. acc.
                               &98.94	 &88.32	 &/	    &52.22   &54.97\\
                               \cmidrule(l){2-7} 
\multicolumn{1}{l}{}           & num    
                               &3120    &3120   &1948  &3120    &3120\\ \bottomrule
\end{tabular}
}
\label{table:chestxray}
\end{table}

\subsection{Spine X-ray}
\begin{table}[H]
\centering
\caption{Results of different models on Spine X-ray in ProbMed. The best-performing model in each question category is \textbf{in-bold}, and the second best is \underline{underlined}.}
\resizebox{\linewidth}{!}{
\begin{tabular}{@{}ccccccc@{}}
\toprule
& & \multicolumn{2}{c}{General Question} & \multicolumn{3}{c}{Specialized Question} \\
\cmidrule(r){3-4}
\cmidrule(r){5-7}
& & Modality & Organ & Abnormality & Condition/Finding & Position \\ 
\midrule
\multirow{1}{*}{\textcolor{gray}{Random Choice}}   
& \textcolor{gray}{Acc. with adv. pairs} & \textcolor{gray}{25} & \textcolor{gray}{25} & \textcolor{gray}{50} & \textcolor{gray}{30.95} & \textcolor{gray}{30.99}\\
\midrule 
\multirow{2}{*}{LLaVA-v1}   & Acc. with adv. pairs               
                               &44.55   &45.04  &50    &0.49    &0\\
                               & Avg. acc.
                               &72.27	 &71.78	 &/	    &47.32   &49.42\\
\multirow{2}{*}{LLaVA-v1.6} & Acc. with adv. pairs               
                               &4.45	 &\textbf{82.67}	 &33.33	&1.48    &0.57\\
                               & Avg. acc.
                               &52.22   &90.84	 &/	    &35.87   &42.02\\
\multirow{2}{*}{MiniGPT-v2}    & Acc. with adv. pairs               
                               &2.97    &52.47	 &58.33	&16.33   &4.02\\
                               & Avg. acc.
                               &51.48	 &71.78	 &/	    &53.84   &51.07\\
\midrule                               
\multirow{2}{*}{LLaVA-Med-v1}  & Acc. with adv. pairs               
                               &8.41	 &7.92	 &33.33	&17.82   &5.74\\
                               & Avg. acc.
                               &52.72 	 &28.96	 &/	    &52.82   &47.58\\
\multirow{2}{*}{LLaVA-Med-v1.5}& Acc. with adv. pairs               
                               &46.53	 &71.78	 &50.00	&14.85   &13.32\\
                               & Avg. acc.
                               &73.01 	 &85.89	 &/	    &55.78   &54.79\\ 
\multirow{2}{*}{BiomedGPT}  & Acc. with adv. pairs               
                               &40.09	 &16.83	 &58.33 &12.37   &2.87\\
                               & Avg. acc.
                               &68.06	 &55.19	 &/	    &50.27   &40.77\\                                
\multirow{2}{*}{Med-Flamingo}  & Acc. with adv. pairs               
                               &14.35	 &25.24	 &50 &14.85   &5.17\\
                               & Avg. acc.
                               &57.17	 &62.12	 &/	    &51.09   &48.38\\
\multirow{2}{*}{CheXagent}     & Acc. with adv. pairs               
                               &82.17	 &20.29	 &\underline{62.5}  &16.83   &0.57\\
                               & Avg. acc.
                               &91.08   &50.74	 &/	    &52.70   &48.70\\
\midrule 
\multirow{2}{*}{GPT-4o}        & Acc. with adv. pairs               
                               &\textbf{95.54}	 &\textbf{79.70}	 &47.82	&\textbf{34.15}   &\textbf{25.86}\\
                               & Avg. acc.
                               &97.02	 &89.60	 &/	    &68.99   &66.03\\
\multirow{2}{*}{GPT-4V}        & Acc. with adv. pairs               
                               &85.57	 &\underline{72.13}	 &47.82	&\underline{29.85}   &18.49\\
                               & Avg. acc.
                               &92.03	 &85.32	 &/	    &65.20   &57.18\\
\multirow{2}{*}{Gemini Pro}    & Acc. with adv. pairs               
                               &\underline{95.02}	 &70.14	 &\textbf{70.83}	&17.91   &\underline{19.07}\\
                               & Avg. acc.
                               &96.76	 &84.82	 &/	    &58.04    &61.72\\
                               \cmidrule(l){2-7} 
\multicolumn{1}{l}{}           & num    
                               &201     &201    &24    &201     &201\\ \bottomrule
\end{tabular}
}
\end{table}

\subsection{Abdominal X-ray}
\begin{table}[H]
\centering
\caption{Results of different models on Abdominal X-ray in ProbMed. The best-performing model in each question category is \textbf{in-bold}, and the second best is \underline{underlined}.}
\resizebox{\linewidth}{!}{
\begin{tabular}{@{}ccccccc@{}}
\toprule
& & \multicolumn{2}{c}{General Question} & \multicolumn{3}{c}{Specialized Question} \\
\cmidrule(r){3-4}
\cmidrule(r){5-7}
& & Modality & Organ & Abnormality & Condition/Finding & Position \\ 
\midrule
\multirow{1}{*}{\textcolor{gray}{Random Choice}}   
& \textcolor{gray}{Acc. with adv. pairs} & \textcolor{gray}{25} & \textcolor{gray}{25} & \textcolor{gray}{50} & \textcolor{gray}{36.55} & \textcolor{gray}{37.46}\\
\midrule 
\multirow{2}{*}{LLaVA-v1}   & Acc. with adv. pairs               
                               &53.87   &53.01  &50    &2.15    &0.56\\
                               & Avg. acc.
                               &76.93	 &76.50	 &/	    &49.14   &50.00\\
\multirow{2}{*}{LLaVA-v1.6} & Acc. with adv. pairs               
                               &5.17	 &56.46	 &46	&6.46    &1.12\\
                               & Avg. acc.
                               &52.15   &75.64	 &/	    &47.63   &48.16\\
\multirow{2}{*}{MiniGPT-v2}    & Acc. with adv. pairs               
                               &4.74    &38.79	 &50	&18.53   &5.64\\
                               & Avg. acc.
                               &52.37	 &67.24	 &/	    &53.65   &50.23\\
\midrule                               
\multirow{2}{*}{LLaVA-Med-v1}  & Acc. with adv. pairs               
                               &7.75	 &42.24	 &60	&14.65   &4.51\\
                               & Avg. acc.
                               &53.23 	 &68.96	 &/	    &47.47   &50.87\\
\multirow{2}{*}{LLaVA-Med-v1.5}& Acc. with adv. pairs               
                               &52.58	 &50.86	 &50.00	&6.46    &14.68\\
                               & Avg. acc.
                               &76.07 	 &73.49	 &/	    &52.02   &54.75\\
\multirow{2}{*}{BiomedGPT}  & Acc. with adv. pairs               
                               &35.77	 &1.29	 &50.00 &10.34   &4.51\\
                               & Avg. acc.
                               &65.30	 &37.50	 &/	    &52.94   &46.79\\                                
\multirow{2}{*}{Med-Flamingo}  & Acc. with adv. pairs               
                               &28.01	 &34.48	 &50    &14.65   &4.51\\
                               & Avg. acc.
                               &64.00	 &66.37	 &/	    &52.52   &46.25\\                               
\multirow{2}{*}{CheXagent}     & Acc. with adv. pairs               
                               &77.15	 &23.70	 &\underline{70}    &12.93   &2.25\\
                               & Avg. acc.
                               &88.57   &52.80	 &/	    &51.30   &49.64\\
\midrule  
\multirow{2}{*}{GPT-4o}        & Acc. with adv. pairs               
                               &\textbf{98.26}	 &\underline{61.47}	 &\underline{70}	&\underline{27.27}   &\underline{21.46}\\
                               & Avg. acc.
                               &99.13	 &79.22	 &/	    &61.83   &59.81\\
\multirow{2}{*}{GPT-4V}        & Acc. with adv. pairs               
                               &84.84	 &50.21	 &60	&\textbf{31.16}   &\textbf{23.16}\\
                               & Avg. acc.
                               &92.42	 &71.42	 &/	    &59.63   &57.03\\
\multirow{2}{*}{Gemini Pro}    & Acc. with adv. pairs               
                               &\underline{97.14}	 &\textbf{63.36}	 &\textbf{85}	&27.15   &19.2\\
                               & Avg. acc.
                               &98.70	 &80.81	 &/	    &59.97   &58.80\\
                               \cmidrule(l){2-7} 
\multicolumn{1}{l}{}           & num    
                               &232     &232    &20    &232     &232\\ \bottomrule
\end{tabular}
}
\end{table}

\section{Dataset Statistics}
\label{appendix:statistics}
In this section, we detail the dataset statistics.

Table~\ref{table:question_per_image} shows the number of questions across each question type. For each image sample, ground-truth questions were created based on available metadata with "yes" answers. For each ground-truth question, we also created a corresponding adversarial question by selecting random adversarial entities and assigning "no" answers. 
For an image showing a normal organ without abnormality, since there is no ground-truth information on the existence of the condition and position, we only construct hallucinated questions for the condition/finding question type. For an image showing abnormality, the number of question pairs per category equals the number of existing conditions or positions.

\begin{table}[H]
\centering
\caption{Number of questions across each question type for each image. }
\resizebox{\linewidth}{!}{
\begin{tabular}{lcc}
\toprule
Question type & 
\begin{tabular}[c]{@{}l@{}}Image with\\ Normal Organ\end{tabular} & 
\begin{tabular}[c]{@{}l@{}}Image with \\ Abnormality\end{tabular} \\ 
\midrule
Modality          & 2 & 2                        \\
Organ             & 2 & 2                        \\
Abnormality       & 1 & 1                        \\
Condition/Finding & 1 & 2 x number of existing conditions \\
Position          & 0 & 2 x number of existing positions   \\ 
\bottomrule
\end{tabular}
}
\label{table:question_per_image}
\end{table}

\section{Robustness Study}
\label{appendix:robustness}

We conducted an additional robustness analysis using GPT-4o---the best-performing model from our evaluations---on the subset of 100 images and 1,090 VQA pairs used in our expert validation. 
The following table presents results on randomly sampled test samples from the expert-validated subset (20\%, 40\%, 60\%, 80\%, and 100\%) and compares accuracy across all samples versus manually validated ("valid") samples:

\begin{table}[H]
\centering
\caption{Valid percentage and accuracy under all samples vs. valid samples on different test sample splits. $N_i$ denotes the number of images in the split, and $N$ denotes the number of corresponding question-answer pairs.}
\resizebox{\linewidth}{!}{
\begin{tabular}{lccc}
\toprule
$N_i$, $N$ & Valid QA (\%) & Acc. all samples (\%) & Acc. valid samples (\%) \\
\midrule
20, 218     & 100.00 & 52.63 & 52.63 \\
40, 421     & 96.19  & 54.36 & 55.19 \\
60, 596     & 96.81  & 55.32 & 55.25 \\
80, 821     & 95.73  & 54.21 & 54.55 \\
100, 1090   & 97.79  & 55.63 & 55.59 \\
6303, 57132 & /      & 55.60 & /     \\
\bottomrule
\end{tabular}
}
\label{table:qa_accuracy}
\end{table}

From these results, we observe minimal differences between the accuracy of all samples versus the validated subset. This indicates our evaluation framework is highly robust to the approximately 6\% error rate identified during pre-processing. Furthermore, the accuracy of GPT-4o on the full ProbMed dataset (55.60\%) closely aligns with the performance on the subset (55.63\%), confirming the generalizability of our robustness findings.

\section{Impact of Chain-of-Thought Prompts and Visual Descriptions on Model Performance}
\label{appendix:ablation_breakdown_results}

In this section, we illustrate model performance under different categories under the three prompting settings: Vanilla, Chain-of-Thought, and Chain-of-Thought with GPT-4o Visual Understanding.

\begin{figure}[htbp]
    \centering
    \includegraphics[width=0.7\linewidth]{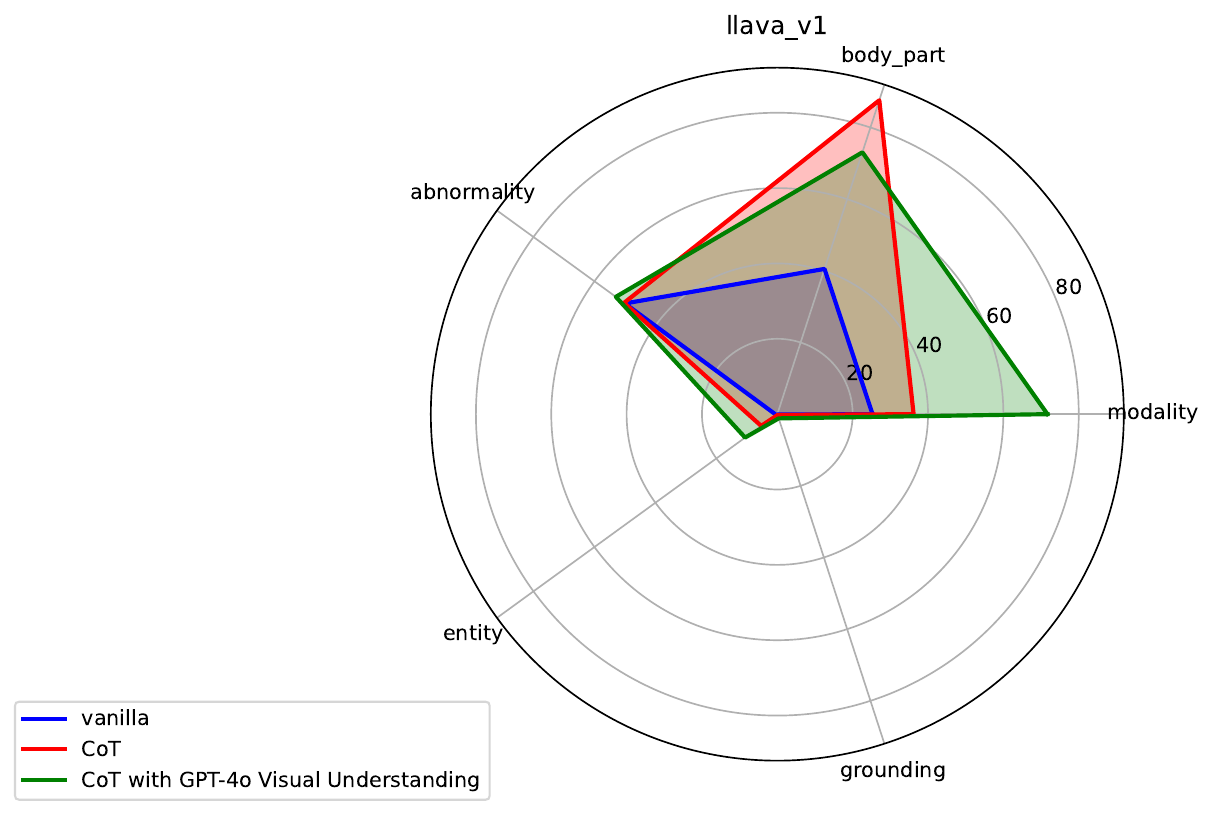}
    \caption{Accuracy of the LLaVA-v1 model across five diagnostic categories under three settings: vanilla (blue), chain-of-thought (CoT, red), and CoT with GPT-4o Visual Understanding (green).
    }
\end{figure}

\begin{figure}[htbp]
    \centering
    \includegraphics[width=0.7\linewidth]{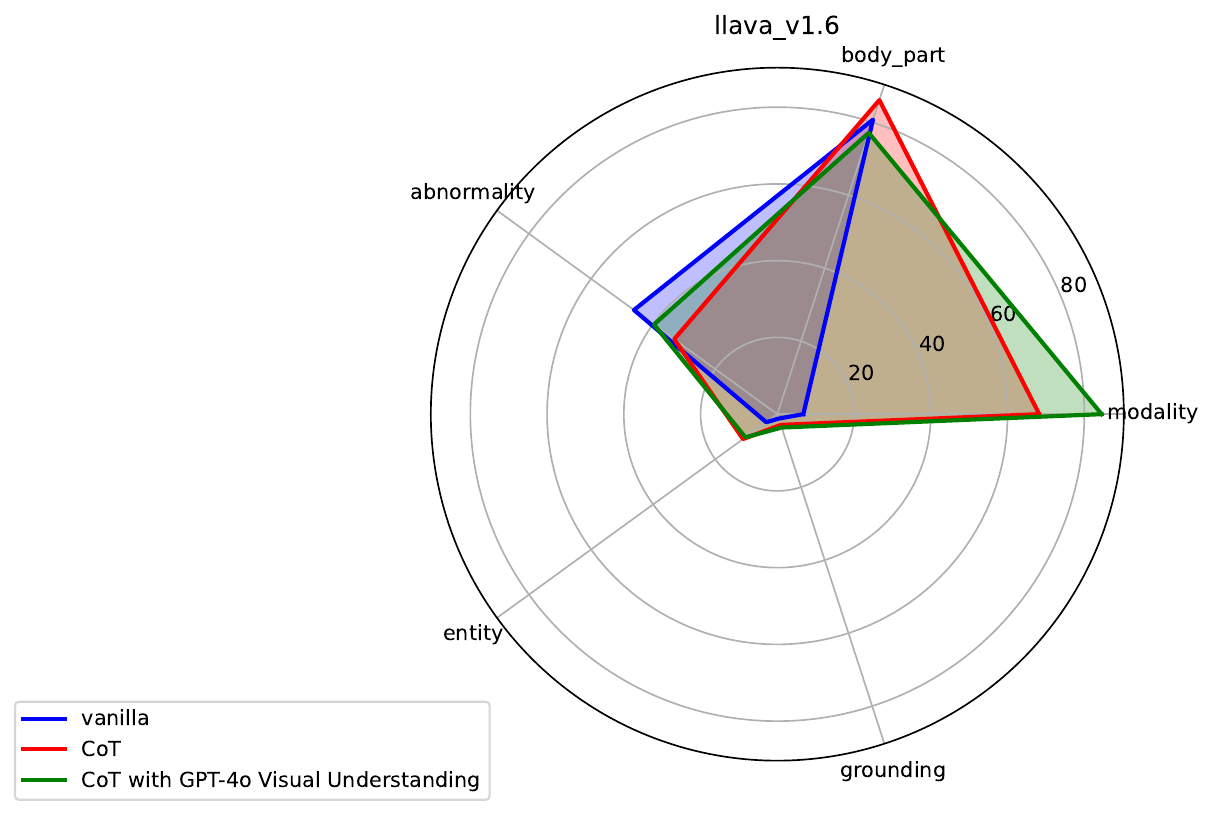}
    \caption{Accuracy of the LLaVA-v1.6 model across five diagnostic categories under three settings: vanilla (blue), chain-of-thought (CoT, red), and CoT with GPT-4o Visual Understanding (green).
    }
\end{figure}

\begin{figure}[htbp]
    \centering
    \includegraphics[width=0.7\linewidth]{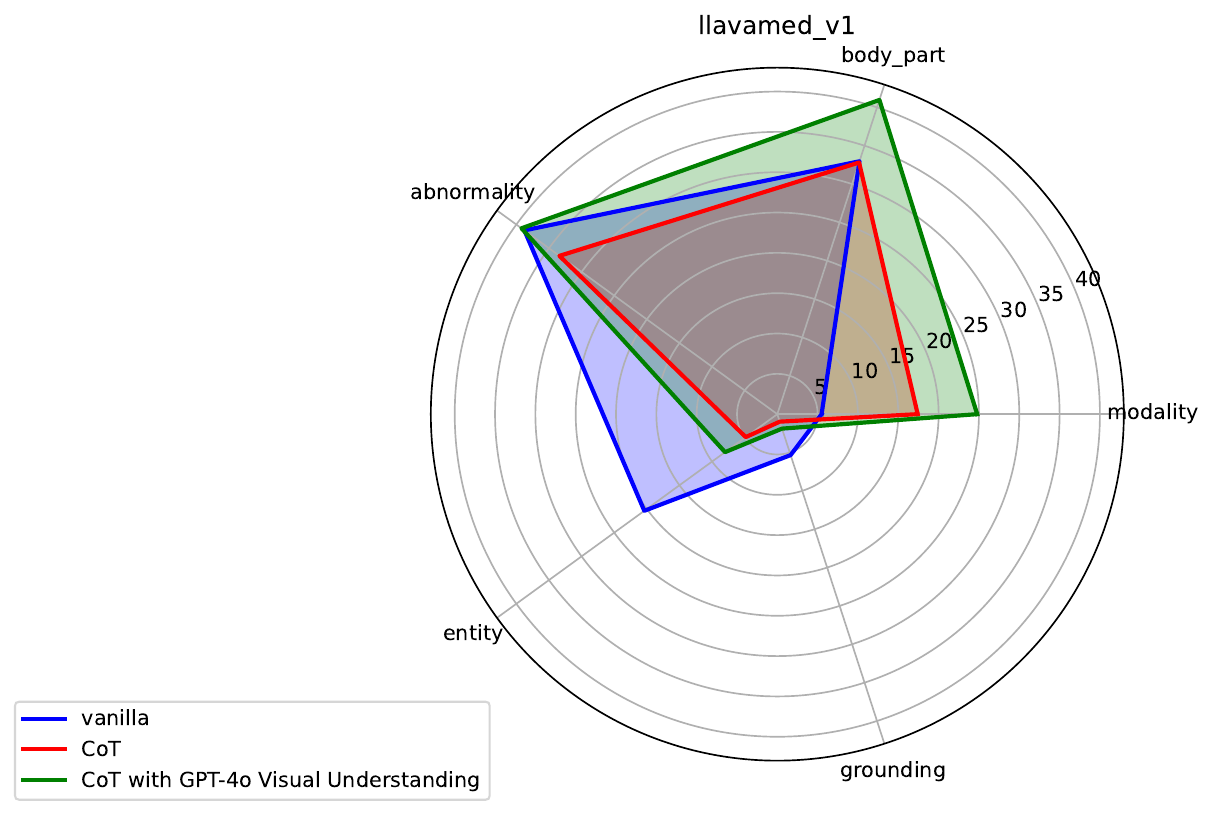}
    \caption{Accuracy of the LLaVA-Med-v1 model across five diagnostic categories under three settings: vanilla (blue), chain-of-thought (CoT, red), and CoT with GPT-4o Visual Understanding (green).
    }
\end{figure}

\begin{figure}[htbp]
    \centering
    \includegraphics[width=0.7\linewidth]{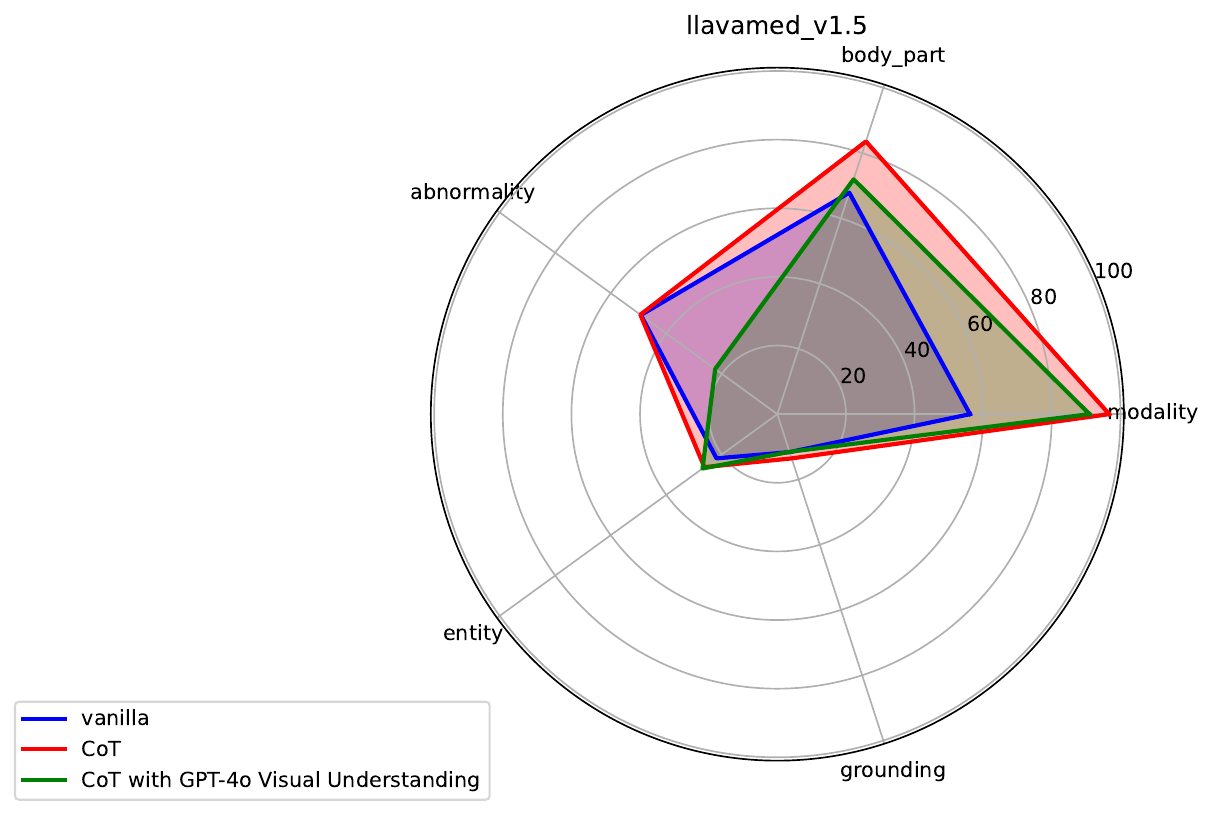}
    \caption{Accuracy of the LLaVA-Med-v1.5 model across five diagnostic categories under three settings: vanilla (blue), chain-of-thought (CoT, red), and CoT with GPT-4o Visual Understanding (green).
    }
\end{figure}

\begin{figure}[htbp]
    \centering
    \includegraphics[width=0.7\linewidth]{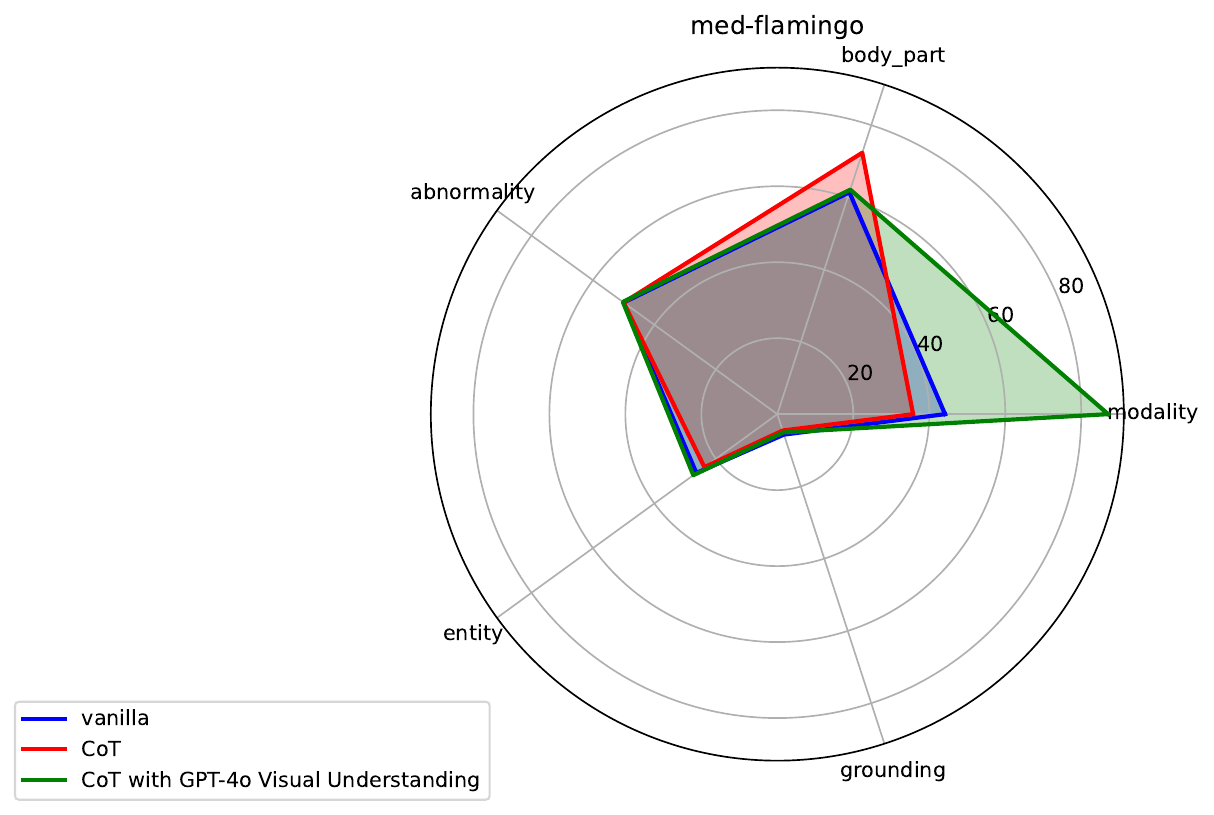}
    \caption{Accuracy of the Med-Flamingo model across five diagnostic categories under three settings: vanilla (blue), chain-of-thought (CoT, red), and CoT with GPT-4o Visual Understanding (green).
    }
\end{figure}

\begin{figure}[htbp]
    \centering
    \includegraphics[width=0.7\linewidth]{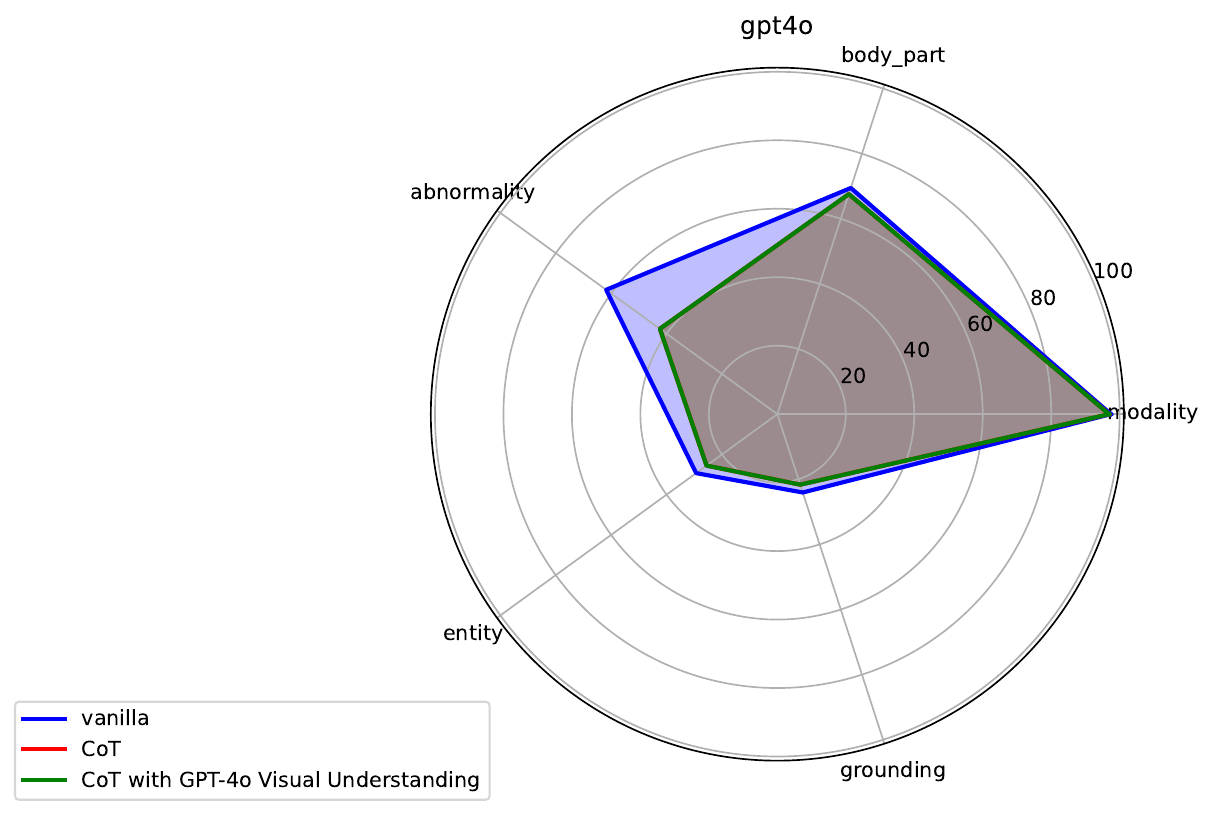}
    \caption{Accuracy of the GPT-4o model across five diagnostic categories under three settings: vanilla (blue), chain-of-thought (CoT, red), and CoT with GPT-4o Visual Understanding (green).
    }
\end{figure}

\newpage
\section{Prompt Details}
\label{appendix:prompt_details}
Below is the few-shot model prompt used for GPT-4 to extract medical conditions and their locations from image captions:

\lstset{
    basicstyle=\tiny\ttfamily,
    breaklines=true,
}
\begin{lstlisting}
You are a helpful assistant and you are given a caption describing a medical image. Extract medical conditions and diseases, along with their locations, if specified. Do not include any information that cannot be directly inferred from the image, for example, patient status or patient history. Outputs should be in the format: "<condition/disease1> : <location1>, <condition/disease2> : <location2>...". The term "<location>" should include at least one positional descriptor and should be explicit in the original caption along with the condition/disease. Otherwise, it should be replaced with "None".

For example, consider the caption: "Fig. 1. MRI abdomen and pelvis showing the cervical mass." The output should be "<cervical mass> : None". For the caption: "Chest radiograph shows enlargement of the hilar mass with spread into the left lower lobe." The output should be "<enlargement of the hilar mass> : <left lower lobe>". Similarly, for the caption: "Abdominal CT scan reveals an enhancing rounded pseudo-aneurysm in the cystic artery, alongside high-density material within the gallbladder's lumen and near the gastrohepatic ligament." The correct output is "<enhancing rounded pseudo-aneurysm> : <cystic artery>, <high-density material> : <lumen of the gallbladder and region of the gastrohepatic ligament>".

Make sure that the response contains only the information in the original caption without adding extra details.

\end{lstlisting}

\section{Response Distribution Visualization within each Category}
\label{appendix:distribution}

Below, we illustrate the distribution of model responses within each category, valid "yes" responses are in red, and valid "no" responses are in blue.

\begin{figure}[htbp]
    \centering
    \includegraphics[width=0.6\linewidth]{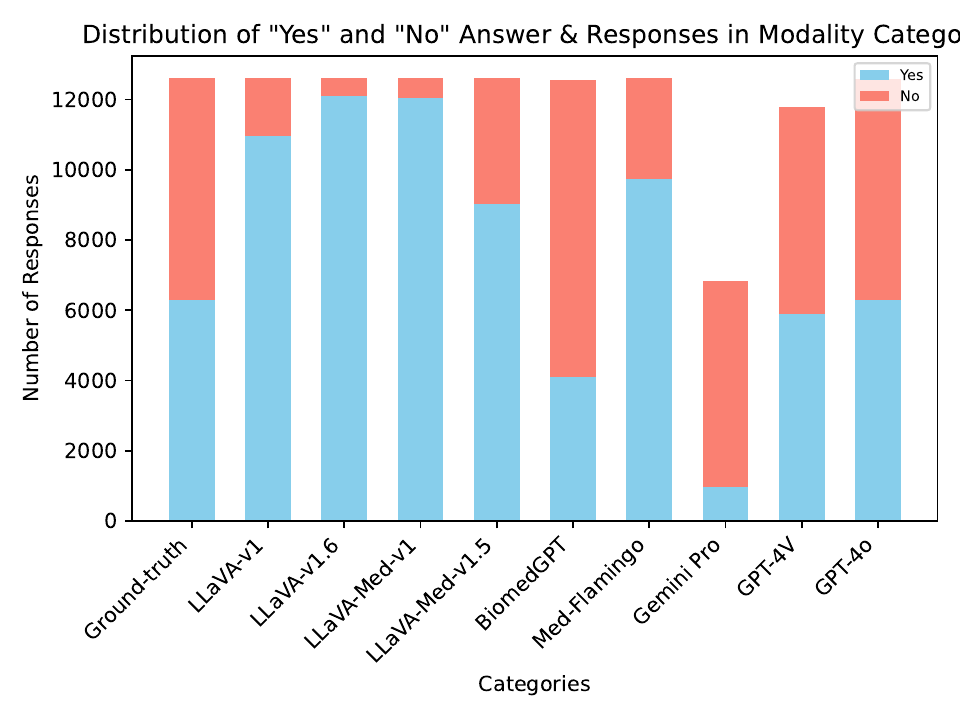}
    \caption{Distribution plot of "yes and "no" ground-truth answers and model responses within the Modality category.}
\end{figure}

\begin{figure}[htbp]
    \centering
    \includegraphics[width=0.6\linewidth]{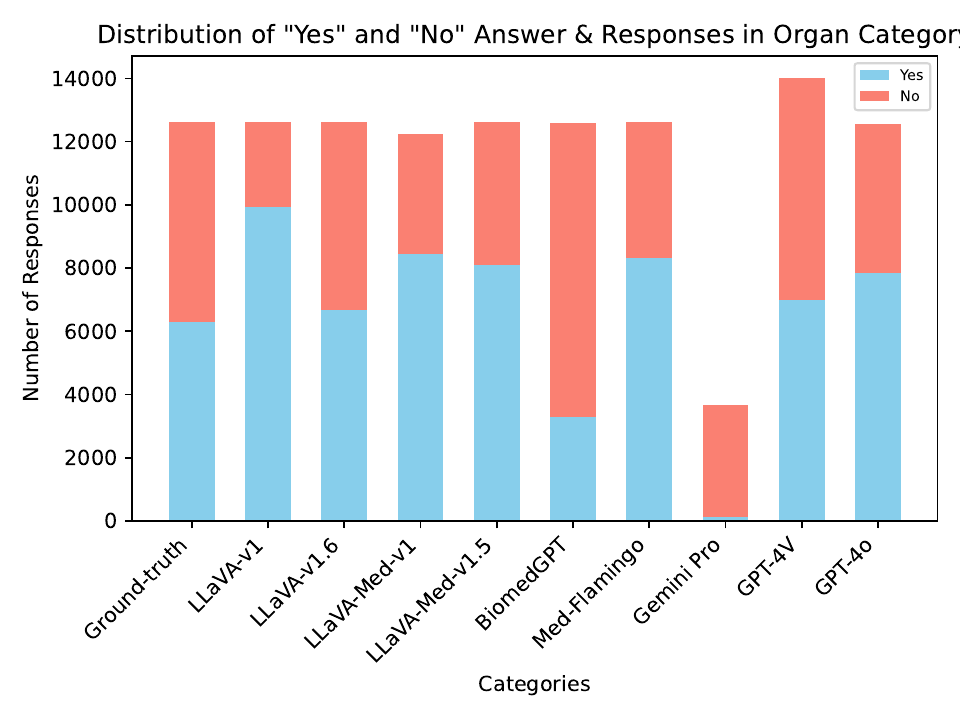}
    \caption{Distribution plot of "yes and "no" ground-truth answers and model responses within the Organ category.}
\end{figure}

\begin{figure}[htbp]
    \centering
    \includegraphics[width=0.6\linewidth]{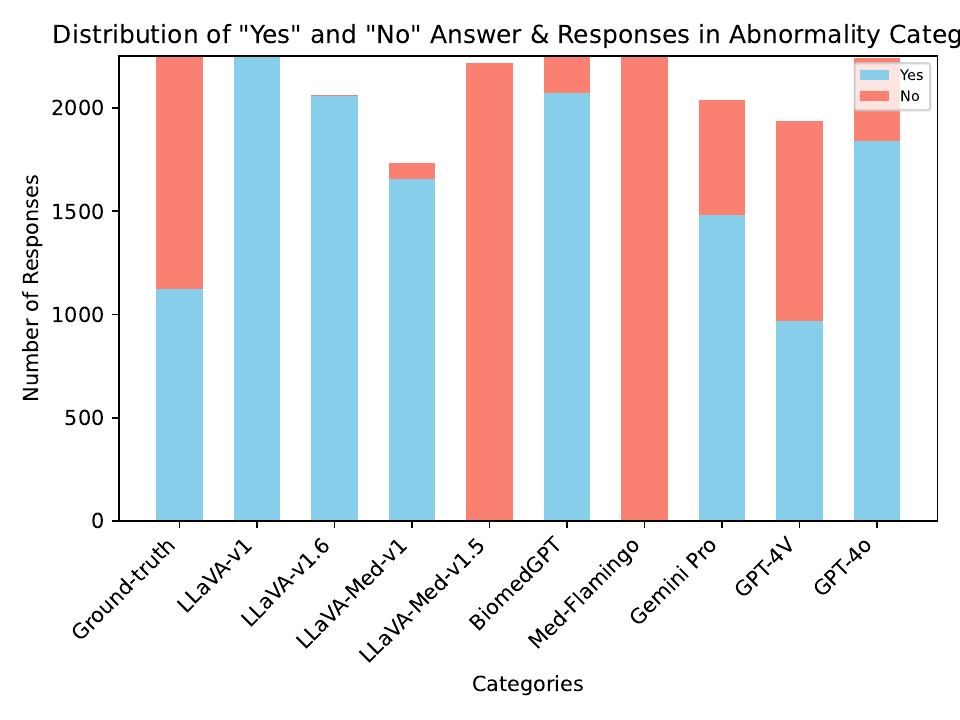}
    \caption{Distribution plot of "yes and "no" ground-truth answers and model responses within the Abnormality category.}
\end{figure}

\begin{figure}[htbp]
    \centering
    \includegraphics[width=0.6\linewidth]{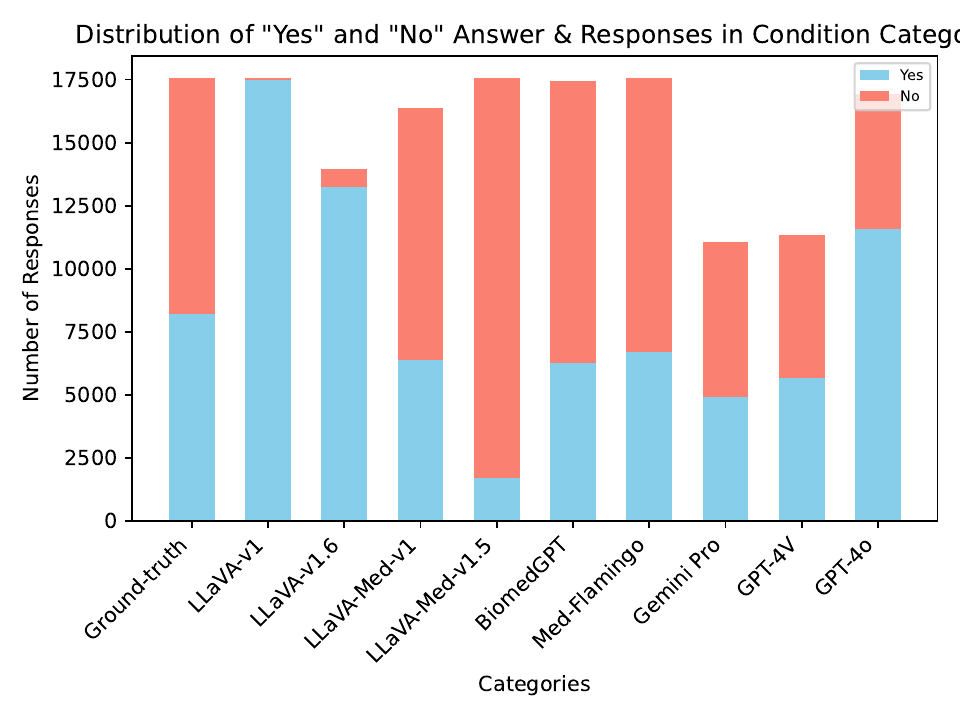}
    \caption{Distribution plot of "yes and "no" ground-truth answers and model responses within the Condition/Finding category.}
\end{figure}

\begin{figure}[htbp]
    \centering
    \includegraphics[width=0.6\linewidth]{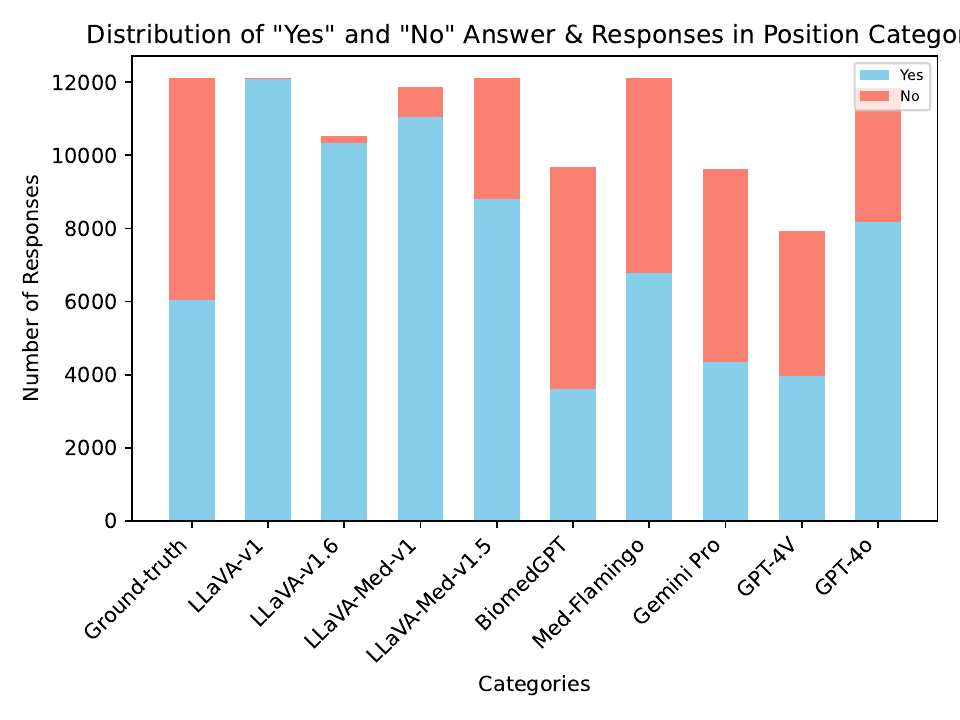}
    \caption{Distribution plot of "yes and "no" ground-truth answers and model responses within the Position category.}
\end{figure}

\end{document}